\documentclass[11pt]{article}
\usepackage{acl}
\usepackage{times}
\usepackage{latexsym}
\usepackage[T1]{fontenc}
\usepackage[utf8]{inputenc}
\usepackage{microtype}
\usepackage{inconsolata}
\usepackage{graphicx}
\usepackage{booktabs}
\usepackage{array}
\usepackage{colortbl}
\usepackage{multirow}
\usepackage{amsmath, amssymb}
\usepackage{hyperref}
\usepackage{url}
\usepackage{listings}
\usepackage{xcolor}
\usepackage{tikz}
\usepackage{subcaption}
\usetikzlibrary{positioning, arrows.meta, calc, shapes.geometric}

\lstdefinelanguage{ProtStructQA}{
  morekeywords={plddt,mean_plddt,mean_pae,max_pae,count_high_pae,distance,ss,
                rel_sasa,mean_rel_sasa,n_neighbors,contact_density,
                radius_of_gyration,n_helices,longest_run,length,size,abs,max,
                argmin,argmax,exists,filter,count,sliding_window,all_pairs,
                all_residues,range,residue,first,last,
                if,then,else,where,by,in,and,or,not,true,false,between},
  morestring=[b]",
  sensitive=true,
  basicstyle=\ttfamily\small,
  keywordstyle=\bfseries\color{blue!60!black},
  stringstyle=\color{green!50!black},
}

\title{ProtStructQA: A Denotation Threshold in Protein Structural Reasoning}

\author{Aravind Mandiga, Guoming Li, Jin Lu, Ismailcem Budak Arpinar, Khaled Rasheed, Samuel E. Aggrey \\
        University of Georgia \\
        \texttt{\{aravind.mandiga, gmli, jin.lu, budak, khaled, saggrey\}@uga.edu}}

\begin{document}
\maketitle
\raggedbottom

\begin{abstract}
Protein-language systems are often evaluated by whether they
generate plausible biological text, but a structural question has
a sharper semantics: it denotes a measurement in a 3D coordinate
system. We introduce \textbf{ProtStructQA}, an executable benchmark
for protein structural question answering in which each
natural-language question is generated from a hidden typed
domain-specific language (DSL) program and the answer is obtained
by executing that program on an AlphaFold-predicted structure.
ProtStructQA releases $382.2$K questions covering confidence,
distances, predicted aligned error (PAE), solvent exposure,
secondary structure, topology and contacts, and held-out
compositions: a $330$K active benchmark over $10$K proteins from
four species, plus a $52.2$K hard-negative robustness pool. Without fine-tuning,
we evaluate Qwen3 models from $0.6$B to $8$B under direct
prompting, chain-of-thought, grammar-constrained executable
voting, executable voting with chain-of-thought, and multi-turn
ReAct-style tool use, and replicate the headline finding on
Gemma-3-$1$B and Gemma-3-$12$B. We find a capability-dependent
\emph{denotation threshold} between Qwen3-$1.7$B and Qwen3-$4$B:
below it,
tool-mediated ReAct dominates because models often fail to produce
executable denotations; above it, chain-of-thought flips from
mostly harmful to strongly beneficial and becomes the strongest
strategy on most splits. Parse-failure and family-level analyses
show that the threshold is a transition from unparseable language
to executable structural denotation, while grammar and execution
remain selectively valuable for PAE and secondary-structure
queries. ProtStructQA reframes scientific QA as compilation from
language to measurement and provides a diagnostic testbed for when
language models can map words to executable 3D structural measurements.
\end{abstract}

\begin{figure}[t!]
\centering
\includegraphics[width=\columnwidth]{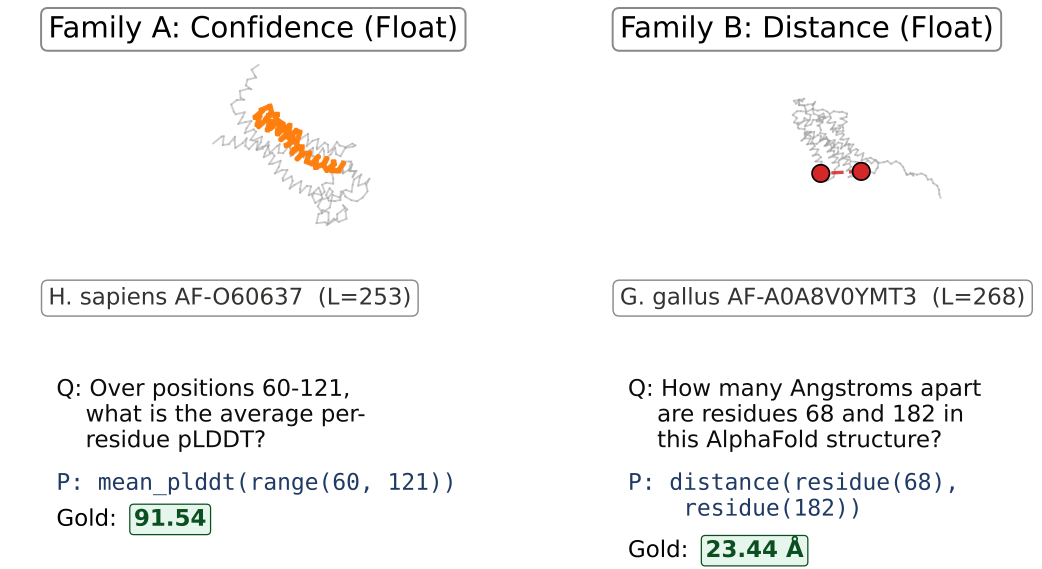}
\caption{\textbf{Two ProtStructQA examples.} Each panel: an
AlphaFold-predicted protein, natural-language question, hidden DSL
program, and gold answer (executed).}
\label{fig:protein_panel}
\end{figure}

\section{Introduction}
\label{sec:intro}

AlphaFold~\citep{jumper2021alphafold} and the AlphaFold Protein
Structure Database (AFDB)~\citep{varadi2022alphafold,bertoni2025afdb}
have changed the bottleneck in structural biology. For a large
fraction of catalogued proteins, a predicted 3D structure is now
available, and AFDB provides open access to more than $200$ million
predicted structures. The remaining challenge is not only to
obtain structures, but to ask precise questions of them: which
region is low-confidence, how far apart are two residues, whether
a window is both high-confidence and contact-rich, or whether two
domains have high predicted aligned error. These are not merely
textual questions. They are measurements over a molecular object.

This creates a mismatch for evaluating large language models
(LLMs). If a model is asked, ``How many Angstroms separate the
alpha-carbons of residues 207 and 220 in the AlphaFold
structure?'', a fluent answer is not enough; the answer should
be correct only if it matches the value computed from the
predicted coordinates. String-overlap
metrics and LLM judges can assess whether a response resembles a
reference, but they cannot by themselves verify that a number,
residue set, region, or structural label was derived from the
protein structure.

We treat protein structural QA as \emph{denotational
semantics}: a natural-language question corresponds to a hidden
typed program, and its meaning is the value obtained by executing
that program on a protein structure. For example, the question above denotes
\texttt{distance(residue(207), residue(220))}; running this
program on the AlphaFold prediction for human protein A6ND36
yields the gold answer ($16.25$~\AA{}). Under this view,
evaluation becomes deterministic: a model is correct only when
its prediction matches the structure-derived denotation.

We introduce \textbf{ProtStructQA}, a large-scale executable
benchmark for residue-level structural QA over AlphaFold-predicted
proteins. ProtStructQA releases $382.2$K questions: a $330$K
active benchmark across seven structural families and a $52.2$K
hard-negative robustness pool. The benchmark covers $10$K proteins
from human, mouse, fruit fly, and chicken, and asks about pLDDT
confidence, $C_\alpha$ distances, PAE, solvent exposure, secondary
structure, contacts, topology, and held-out compositions of these
primitives. Each question is generated from a hidden DSL program
and evaluated by deterministic execution rather than text overlap.

This setup lets us ask a controlled inference-time question:
\emph{where is the structural denotation computed?} Direct
prompting compiles in one step; chain-of-thought plans in
language; grammar-constrained executable voting samples and
executes candidate programs; ReAct-style tool use externalizes
structural access through multi-turn tool calls. A shared
executor isolates whether failures arise from language planning,
program syntax, structural computation, or answer formatting.

Across Qwen3 models from $0.6$B to $8$B, we observe a
capability-dependent \emph{denotation threshold} between
Qwen3-$1.7$B and Qwen3-$4$B, replicated on Gemma-3-$1$B and
Gemma-3-$12$B. Below the threshold, small models often cannot
produce executable denotations, and multi-turn tool use dominates.
Above the threshold, chain-of-thought flips from mostly harmful
to strongly beneficial, and free-form CoT or executed CoT wins
nearly all evaluation cells; the transition is mechanistically a
shift from unparseable to valid structural denotations.

Our contributions are:
\begin{itemize}\setlength\itemsep{1pt}
  \item \textbf{Executable structural QA.}
        We introduce ProtStructQA, a large-scale benchmark in which
        each natural-language protein question has a hidden typed
        DSL program and a gold answer obtained by deterministic
        execution on an AlphaFold-predicted structure.
  \item \textbf{Typed denotational semantics for protein questions.}
        We formalize structural QA as compilation from language to
        a measurement algebra over residues, regions, residue
        pairs, confidence, PAE, solvent exposure, secondary
        structure, and contacts.
  \item \textbf{Controlled inference-time evaluation.}
        We compare direct prompting, chain-of-thought,
        grammar-constrained executable voting, executable voting
        with CoT, and ReAct-style tool use without fine-tuning,
        across Qwen3 ($0.6$B--$8$B) and a Gemma-3 cross-family
        replication ($1$B and $12$B).
  \item \textbf{The denotation threshold.}
        We show a scale-dependent crossover: sub-threshold models
        benefit most from tool-mediated denotation, while
        supra-threshold models use chain-of-thought as an internal
        compiler from language to structural measurement.
\end{itemize}

\section{Related Work}
\label{sec:related}

\paragraph{Protein representation and foundation-model benchmarks.}
TAPE~\citep{rao2019tape}, PEER~\citep{xu2022peer}, and
ProteinGym~\citep{notin2023proteingym} collectively evaluate
protein representation, function/localization/interaction
prediction, and mutation-effect tasks. These suites are essential for
measuring protein foundation models, but they are not designed to evaluate
natural-language questions whose answers must be computed from
residue-level 3D structural measurements. ProtStructQA is
complementary: it evaluates whether a language model can map a
surface question to an executable operation over an
AlphaFold-predicted structure.

\paragraph{Protein-language QA and chat systems.}
Recent work has begun to connect proteins and natural language
through QA datasets, instruction tuning, protein-to-text
generation, and multimodal protein chat.
PQA/Pika~\citep{carrami2024pqa} introduces a curated protein QA
benchmark for free-form enquiry, contrasting with earlier PDB-QA
datasets of predefined questions over PDB entries;
ProtT3~\citep{liu2024prott3} generates text descriptions from
protein-sequence inputs;
Mol-Instructions~\citep{fang2024molinstructions} and
InstructProtein~\citep{wang2024instructprotein} align biomolecular
knowledge with natural-language supervision; and
ProtChatGPT~\citep{wang2024protchatgpt},
ProteinGPT~\citep{xiao2024proteingpt}, and
Prot2Chat~\citep{wang2025prot2chat} let users query protein
sequence and/or structure through LLM interfaces. These efforts
evaluate generation with lexical, semantic, or LLM-judge metrics,
or task-specific QA scoring. ProtStructQA
differs in its evaluation target: each question has a hidden
residue-level DSL program and a deterministic answer produced by
execution over predicted 3D coordinates.

\paragraph{Executable and compositional reasoning benchmarks.}
ProtStructQA inherits from a tradition of compositional reasoning
benchmarks: SCAN~\citep{lake2018scan} (grammar-based compositional
generalization), CFQ~\citep{keysers2020cfq} (executable SPARQL
queries), CLEVR~\citep{johnson2017clevr} (functional programs over
synthetic visual scenes), and GQA~\citep{hudson2019gqa} (scene-graph
question generation with functional programs). ProtStructQA imports this denotational idea into a scientific
domain: the ``scene'' is an AlphaFold-predicted protein, the
objects are residues and regions, and the answer is a structural
measurement rather than a visual attribute or database fact.

\paragraph{Structured decoding, programs, and agents.}
Grammar-constrained decoding enforces output structure at the
token level~\citep{geng2023grammar,willard2023outlines,koo2024automataaugmented,park2025flexiblegcd},
and JSONSchemaBench~\citep{geng2025jsonschema} benchmarks
syntactic validity at scale. Program-aided methods such
as Program-of-Thought~\citep{chen2023program} and PAL~\citep{gao2023pal}
separate reasoning from computation;
self-consistency~\citep{wang2023selfconsistency} aggregates
multiple reasoning traces. ReAct~\citep{yao2023react} interleaves
reasoning with tool use. ProtStructQA uses these
methods not merely to test format compliance, but to ask when
valid structure becomes correct scientific denotation: a
syntactically valid DSL program is useful only when its execution
matches the structural quantity requested by the question.

\paragraph{Positioning.}
In summary, ProtStructQA is not simply another protein QA dataset
or another constrained-decoding benchmark. It combines protein
structural data, typed executable semantics, and controlled
inference-time reasoning to evaluate whether language models can
compile natural-language questions into measurements over predicted
molecular structure. This intersection is the paper's contribution.

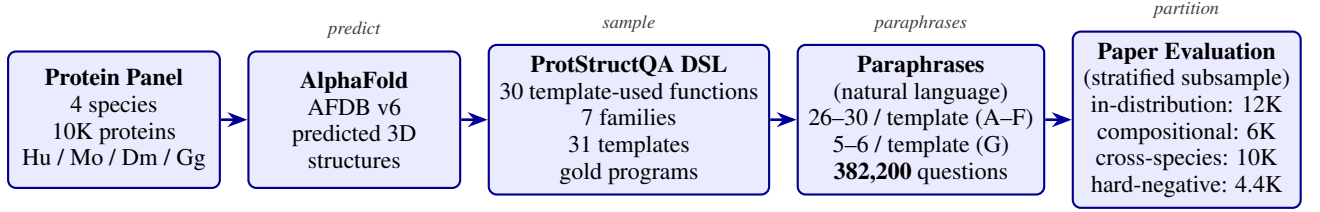
\begin{figure*}[t]
\centering
\begin{tikzpicture}[
  every node/.style={font=\small},
  stage/.style={rectangle, draw=blue!55!black, thick, rounded corners=3pt,
                minimum width=2.8cm, minimum height=1.8cm, align=center,
                fill=blue!8, inner sep=4pt, font=\footnotesize},
  arrow/.style={-Stealth, thick, blue!60!black, line width=1pt},
  ann/.style={font=\scriptsize\itshape, gray!50!black},
]
\node[stage] (s1) {%
  \textbf{Protein Panel} \\
  4 species \\
  10K proteins \\
  Hu / Mo / Dm / Gg};
\node[stage, right=0.35cm of s1] (s2) {%
  \textbf{AlphaFold} \\
  AFDB v6 \\
  predicted 3D \\
  structures};
\node[stage, right=0.35cm of s2] (s3) {%
  \textbf{ProtStructQA DSL} \\
  30 template-used functions \\
  7 families \\
  31 templates \\
  gold programs};
\node[stage, right=0.35cm of s3] (s4) {%
  \textbf{Paraphrases} \\
  (natural language) \\
  26--30 / template (A--F) \\
  5--6 / template (G) \\
  \textbf{382{,}200} questions};
\node[stage, right=0.35cm of s4] (s5) {%
  \textbf{Paper Evaluation} \\
  (stratified subsample) \\
  in-distribution: 12K \\
  compositional: 6K \\
  cross-species: 10K \\
  hard-negative: 4.4K};
\draw[arrow] (s1) -- (s2);
\draw[arrow] (s2) -- (s3);
\draw[arrow] (s3) -- (s4);
\draw[arrow] (s4) -- (s5);
\node[ann, above=0.03cm of s2.north] {predict};
\node[ann, above=0.03cm of s3.north] {sample};
\node[ann, above=0.03cm of s4.north] {paraphrases};
\node[ann, above=0.03cm of s5.north] {partition};
\end{tikzpicture}
\caption{\textbf{ProtStructQA construction pipeline:} from
protein panel through DSL and paraphrases to train-disjoint
evaluation tracks. Pipeline is deterministic given fixed seeds.}
\label{fig:pipeline}
\end{figure*}

\section{The ProtStructQA Benchmark}
\label{sec:benchmark}

ProtStructQA is built around four objects: a predicted structural
world, a natural-language question, a hidden typed denotation, and
a deterministic oracle. The world is an AlphaFold-predicted
protein structure; the question is a paraphrase of a hidden
template; the denotation is a DSL program; and the oracle is the
executor that computes the answer from coordinates, confidence
scores, PAE, solvent exposure, and secondary-structure
annotations. Figure~\ref{fig:pipeline} summarizes the construction
pipeline. In plain terms, each question is generated from a hidden
program, and the correct answer is the value of running that
program on the protein's predicted structure.

\subsection{Protein Panel}
\label{sec:benchmark:panel}

We sample $10{,}000$ AlphaFold-predicted protein structures across
four UniProt reference proteomes~\citep{uniprot2025}:
$4{,}000$ human, $2{,}500$ mouse, $1{,}500$ fruit fly, and
$2{,}000$ chicken (Table~\ref{tab:protein_panel},
App.~\ref{app:errors}). Human serves as the in-distribution
anchor, while the other species define species-stratified
cross-proteome shifts: mouse provides a close mammalian
comparison, chicken a non-mammalian vertebrate comparison, and
fruit fly an invertebrate comparison. The species
selection is intended to create controlled proteome-level
distribution shifts, not to prove broad phylogenetic
generalization. Sampling is length-stratified within species
(final-panel lengths range from $16$ to $2{,}321$ amino acids),
and predicted structures come from AFDB
v6~\citep{bertoni2025afdb}, produced by
AlphaFold~\citep{jumper2021alphafold} and released under
CC~BY~4.0.

For each protein, we extract the $C_\alpha$ trace, per-residue
pLDDT, the PAE matrix, secondary structure assigned by DSSP and
collapsed to helix/strand/coil~\citep{kabsch1983dssp}, and
relative solvent-accessible surface area (SASA). These features define the executor's structural
state, and the benchmark therefore evaluates answers relative to
AlphaFold-predicted structures and the per-residue annotations
derived from them, not experimental structural ground truth.
Figure~\ref{fig:protein_panel} shows two example questions; the
per-species protein counts (Table~\ref{tab:protein_panel}) and the
panel's length, pLDDT, and secondary-structure composition
(Figure~\ref{fig:protein_panel_dist}) are reported in
App.~\ref{app:errors}.

\subsection{Formal denotational setup}
\label{sec:benchmark:formal}

Informally, each question is paired with a hidden program; running
the program on a protein returns a number, a yes/no, a residue, a
region, a residue set, a residue-pair set, or a secondary-structure
label. The formal setup below names the pieces of this pipeline so
later sections can reference them precisely.

Let protein $p$ have structural state
\[
  S_p = (X_p,\, c_p,\, A_p,\, \sigma_p,\, \rho_p),
\]
where
$X_p \in \mathbb{R}^{n_p \times 3}$ are $C_\alpha$ coordinates,
$c_p \in [0, 100]^{n_p}$ are pLDDT scores,
$A_p \in \mathbb{R}_{\ge 0}^{n_p \times n_p}$ is the PAE matrix,
$\sigma_p \in \{\mathtt{H},\mathtt{E},\mathtt{C}\}^{n_p}$ gives
per-residue secondary structure, and
$\rho_p \in [0, 1]^{n_p}$ is relative solvent accessibility. Each
benchmark example is a tuple
\[
  e_i = (p_i,\, q_i,\, z_i,\, y_i,\, \tau_i),
\]
where $q_i$ is the natural-language question, $z_i \in \mathcal{G}$
is the hidden DSL program (\S\ref{sec:benchmark:dsl}), $\tau_i$ is
the answer type, and
\[
  y_i \;=\; E(z_i,\, S_{p_i})
\]
is the gold answer obtained by deterministic execution. Multiple
paraphrases of the same template-parameter assignment map to the
same hidden program and the same denotation, while semantically
similar questions can have different programs. A prediction
$\hat y$ is correct only if it matches $y_i$ under the type-specific
metric defined in \S\ref{sec:experiments}.

\subsection{DSL for structural reasoning}
\label{sec:benchmark:dsl}

The DSL is intentionally small: it is not a full language for
structural biology, but a closed measurement algebra over
residues, regions, residue pairs, and proteins, covering common
structural operations while keeping every question executable and
type-checkable. The DSL operates over typed values
\texttt{Residue}, \texttt{Region}, \texttt{ResidueSet},
\texttt{PairSet}, \texttt{Bool}, \texttt{Int}, \texttt{Float}, and
\texttt{SecStruct}; typed execution rejects ill-formed programs
such as averaging a residue pair or comparing a secondary-structure
label to a distance threshold. Regions are $1$-indexed inclusive
(\texttt{range(s, e)} covers $e{-}s{+}1$ residues).
Primitives cover per-residue
properties (pLDDT, DSSP label, relative SASA, neighbor count),
pairwise properties ($C_\alpha$ distance and PAE, with optional
sequence-separation constraints via \texttt{min\_sep}), and
region- or protein-level aggregates (mean pLDDT, contact density,
radius of gyration, helix counts).
They compose through comparison and Boolean logic with
\texttt{exists}, \texttt{count}, \texttt{filter}, \texttt{argmin},
\texttt{argmax}, sliding windows, and all-pairs quantification,
producing both atomic measurements and held-out compositions (e.g., whether any $40$-residue window is
simultaneously high-confidence and contact-rich). Several
primitives correspond to closed-form structural quantities:
\begin{align*}
  \operatorname{distance}(i, j)
    &= \lVert X_i - X_j \rVert_2, \\
  \operatorname{mean\_plddt}(R)
    &= \tfrac{1}{|R|}\sum_{i \in R} c_i, \\
  \operatorname{mean\_pae}(R_1, R_2)
    &= \tfrac{1}{|R_1||R_2|}\!\!\sum_{i \in R_1, j \in R_2}\!\! A_{ij}, \\
  \operatorname{contact\_density}(R)
    &= \tfrac{1}{\binom{|R|}{2}}
       \!\!\sum_{\substack{i, j \in R \\ i < j}}\!\!
       \mathbf{1}\bigl\{\lVert X_i - X_j \rVert_2 < 8\,\text{\AA}\bigr\}.
\end{align*}
The full $30$-primitive vocabulary, the formal grammar, and the
program pattern of every one of the $31$ templates are listed in
App.~\ref{app:dsl_catalogue}.

\subsection{Question Templates}
\label{sec:benchmark:templates}

Questions span $7$ families and $31$ templates (per-family and
per-template counts in App.~\ref{app:per_template}). Families
A--F cover atomic structural
operations (confidence, distance, PAE, solvent exposure/packing,
secondary structure, contacts/topology).

Each template is rendered through multiple natural-language
paraphrases (median $29$ per A--F template, $5$--$6$ per G
template) spanning terse, interrogative, imperative, and
structural-biology wordings. A paraphrase changes the wording
but not the denotation: for a fixed template and parameter
assignment, all paraphrases map to the same hidden program and
answer. The paraphrase pool was generated by an LLM (Claude
Opus~4.7~\citep{anthropic2026opus47})
and then author-verified to remove paraphrases that revealed the
answer and to make sure every paraphrase kept the same slot names
as its template; consistency between paraphrase and program is then validated
on a 3{,}500-question stratified sample (\S\ref{sec:benchmark:validation}). Family G is not new biology; it is new composition. This is the
closest analogue to compositional splits in SCAN, CFQ, CLEVR, and
GQA, grounded in protein structural measurements
(App.~\ref{app:errors}, Figure~\ref{fig:examples}).

\subsection{Hard Negatives}
\label{sec:benchmark:hn}

Hard negatives are semantic stress tests in structural space.
They preserve much of the linguistic surface while changing the
denotation or crossing a decision boundary, so that template
priors alone are insufficient. They test whether a model binds
the question to the measured structural property, rather than
exploiting wording-only heuristics.

We construct hard negatives by perturbing template parameters
(residue indices, region endpoints, or numeric thresholds) so
that the resulting question is close to an original but has a
different executed answer (two construction types HN1/HN2 with
per-class breakdown in App.~\ref{app:hn_breakdown}). We release
a $52.2$K hard-negative robustness pool and evaluate a stratified
$4{,}357$-question hard-negative split.

\subsection{Generalization axes}
\label{sec:benchmark:splits}

ProtStructQA evaluates four generalization axes
(Table~\ref{tab:split_inventory}). The in-distribution split
($12{,}000$ questions) tests new proteins and parameter values
under seen template families. The compositional split
($6{,}000$ questions, Family G) tests held-out compositions of
known operators. The cross-species split ($10{,}000$ questions)
tests species-stratified cross-proteome shift on mouse, fly, and
chicken proteins. The hard-negative split ($4{,}357$ questions,
\S\ref{sec:benchmark:hn}) tests semantic robustness under
near-surface-preserving structural perturbations.

Within the human in-distribution pool, we partition proteins
$80\%$ training / $10\%$ development / $10\%$ test, with no
protein appearing in more than one split. Family G is held out of
the exemplar pool. We do not fine-tune model weights on any
split. The paper evaluates models on a $32{,}357$-question
stratified subsample of the four evaluation tracks
(\S\ref{sec:benchmark:subsample}). We use ``cross-species'' and ``cross-proteome'' descriptively.
The split measures robustness to proteome-level distribution
shift across our four selected species, not general biological
extrapolation to arbitrary species, protein families, or
experimental conditions.

\subsection{Validation}
\label{sec:benchmark:validation}

The main risk in executable benchmark construction is that errors
in program generation, parsing, or structural preprocessing
become false gold labels. We validate ProtStructQA as a trust
stack: schema and type gates, deterministic execution,
paraphrase-program consistency, cross-tool agreement, and
distributional checks on subsampled splits.

First, every generated question passes a six-step sanity gate
(file format, single-split membership, cross-species questions on
non-human proteins, both HN classes labeled correctly, gold
program re-execution, and a question-only baseline confirming
that the wording alone is insufficient). Repeated execution of
the same program on the same structural state always produces
the same answer.

Second, we verify that paraphrases recover the intended program
parameters: from each of the seven families we sample $500$
questions ($3{,}500$ total) and confirm that every numeric
literal in the gold program appears in the natural-language
question and that the question contains at least one
family-specific keyword. All $3{,}500$ pass.

Third, we recompute representative structural primitives with
independent toolchains:
BioPython~\citep{cock2009biopython} for distances, pLDDT, and
Shrake--Rupley SASA; raw AFDB JSON for PAE; biotite
P-SEA~\citep{labesse1997psea,kunzmann2018biotite} for
secondary structure. All agree on evaluated templates
(App.~\ref{app:validation}).

\label{sec:benchmark:subsample}
Finally, for the three subsampled evaluation splits, stratified
sampling preserves relevant marginals: KL divergence is at most
$0.0037$ for every tracked category (family, species, template,
answer type), so subsample trends transfer to the full pool
within statistical noise (App.~\ref{app:subsample}).

\section{Baseline Suite}
\label{sec:method}

We evaluate five inference-time baselines that differ in
\emph{where the structural denotation is computed.} Direct
prompting asks the model to compile the question into an answer in
one step. Chain-of-thought lets the model plan in natural language
before emitting an answer. Executable voting samples grammar-valid
DSL programs, executes them, and aggregates the executed answers
under a type-aware loss. EV+CoT combines language planning with
executable voting. ReAct-style tool use externalizes structural
computation through a multi-turn tool interface. All are
inference-time only (no fine-tuning) and share the same DSL output
schema; they differ only in decoding strategy
(\S\ref{sec:experiments}). Iterative refinement and
deliberate-search methods~\citep{shinn2023reflexion,yao2023tot,madaan2023selfrefine}
are out of scope.

\begin{itemize}\setlength\itemsep{1pt}
  \item \textsc{Standard}~\citep{brown2020gpt3}
        (\emph{direct compiler}):
        few-shot direct prompting at $T{=}0$; one DSL emission per
        question, compiling $q$ into a typed program in a single
        forward pass.
  \item \textsc{CoT}~\citep{wei2022chain}
        (\emph{internal compiler with language planning}):
        prepends a 4-step task-decomposition checklist
        (answer-type $\to$ primitive $\to$ scope $\to$ thresholds);
        same sampling and exemplars as Standard. Vanilla
        zero-shot~\citep{kojima2022zeroshotcot} and PAL-style
        variants~\citep{gao2023pal,chen2023program} are evaluated as
        prompt-robustness checks (App.~\ref{app:prompt_form}).
  \item \textsc{EV}~\citep{wang2023selfconsistency,gao2023pal,chen2023program}
        (\emph{grammar-constrained executable voting}):
        the program-form analogue of self-consistency. We sample
        $k{=}3$ grammar-constrained programs
        $\hat z_1, \ldots, \hat z_K$ from
        $p_\theta(\cdot \mid q,\, z \in \mathcal{G})$ at $T{=}0.7$,
        top-$p{=}0.95$ via
        LLGuidance~\citep{llguidance2024}, execute each
        to obtain candidate denotations
        $\hat y_k = E(\hat z_k,\, S_p)$, and aggregate by
        type-specific minimum-risk decoding~\citep{eikema2020mbr}:
        \[
          \hat y \;=\;
          \arg\min_{u \in \{\hat y_1, \ldots, \hat y_K\}}
          \sum_{k=1}^{K} \ell_\tau(u,\, \hat y_k),
        \]
        where the type-specific loss is
\begin{equation*}
\resizebox{\linewidth}{!}{$
          \ell_\tau(u, v) =
          \begin{cases}
            \mathbf{1}[u \ne v] &
              \tau \in \{\text{Bool}, \text{Int}, \text{SecStruct}\}, \\
            |u - v| &
              \tau = \text{Float}, \\
            1 - \mathrm{IoU}(u, v) &
              \tau \in \{\text{Region}, \text{ResidueSet}, \text{PairSet}\}.
          \end{cases}
$}
\end{equation*}
        These reduce to majority vote, sample median, and
        IoU-medoid respectively. Thus EV is not only
        self-consistency over text; it is voting in the
        executor's answer space.
  \item \textsc{EV+CoT}
        (\emph{language planning combined with executable voting}):
        EV with the CoT prefix applied per sample, the
        self-consistency $+$ CoT
        combination~\citep{wang2023selfconsistency} on executed
        programs rather than free-form text.
  \item \textsc{ReAct}~\citep{yao2023react}
        (\emph{tool-mediated denotation prosthesis}):
        multi-turn agent interleaving reasoning, tool actions, and
        observations (\texttt{<think>}, \texttt{<act>},
        \texttt{<obs>}, \texttt{<answer>} delimiters).
        Five DSL tools (point-residue lookup, residue-pair
        distance, mean PAE over a region, region summary stats, and
        raw DSL execution), four primer examples,
        $\textsc{max\_turns}{=}8$. Tool calls externalize structural
        access; the model need not internally compile the full
        question into a single expression. The final
        \texttt{<answer>} is free-form text (analyzed in
        \S\ref{sec:results:react}).
\end{itemize}

\section{Experimental Setup}
\label{sec:experiments}

\textbf{Models and inference.}
We evaluate \textsc{Qwen3-0.6B}, \textsc{Qwen3-1.7B},
\textsc{Qwen3-4B}, and \textsc{Qwen3-8B}~\citep{qwen3} via
vLLM~\citep{kwon2023vllm}~v0.16.0 on a single H100-80GB in
bfloat16. All baselines receive the same compact protein summary
(length, mean pLDDT, helix and strand counts, secondary-structure
bands, pLDDT bands) and
four train-pool few-shot exemplars per question (families A--F,
target template excluded); this design isolates inference-time
behavior rather than supervised adaptation to ProtStructQA.
EV and EV+CoT: temperature~0.7, top-$p$~0.95, $k{=}3$,
\texttt{max\_tokens}=192, grammar enforcement via
LLGuidance~\citep{llguidance2024}.
Standard and CoT: temperature~0, \texttt{max\_tokens}=384.
ReAct: temperature~0 (greedy, following~\citet{yao2023react}),
\texttt{max\_tokens}=384/turn, $\textsc{max\_turns}=8$.
Overall GPU usage for the experiments is approximately
$150$ hours on H100.

\textbf{Metrics.}
A prediction is scored under a type-specific tolerance,
\begin{equation*}
\resizebox{\columnwidth}{!}{$
M_\tau(\hat y, y) = \begin{cases}
\mathbf{1}\bigl[|\hat y - y| \le \max(0.5,\ 0.05 \cdot \max(|y|,|\hat y|))\bigr] & \tau = \text{Float}, \\
\mathbf{1}\bigl[|\hat y - y| \le \max(2, 0.10|y|)\bigr] & \tau = \text{Int}, \\
\mathbf{1}[\hat y = y] & \tau \in \{\text{Bool}, \text{SecStruct}, \text{Region}\}, \\
\mathbf{1}\bigl[\mathrm{IoU}(\hat y, y) \ge 0.9\bigr] & \tau \in \{\text{ResidueSet}, \text{PairSet}\},
\end{cases}
$}
\end{equation*}
applied in each Float quantity's native unit (\AA{} for
distance/PAE, pLDDT points for confidence, unitless for SASA
fraction and contact density). Marginal accuracy is the mean,
$\operatorname{Acc} = \tfrac{1}{N}\sum_{i=1}^{N} M_{\tau_i}(\hat y_i, y_i)$.
Following~\citet{dror2018hitchhiker} we report bootstrap $95\%$
CIs (1{,}000 resamples~\citep{efron1993bootstrap};
App.~\ref{app:bootstrap_cis}) on marginal accuracies and
McNemar's test~\citep{mcnemar1947} on
matched-question pairs, with Bonferroni correction tied to the
number of planned comparisons; per-cell statistical significance
for Qwen3 is reported in App.~\ref{app:stats_detail} (Tables
\ref{tab:significance_sub} and \ref{tab:significance_supra}), and
prompt-form robustness in App.~\ref{app:prompt_form}. EV and
EV+CoT run at three seeds; Standard, CoT, and ReAct are reported
single-seed (seed~0).

\section{Results: The Denotation Threshold}
\label{sec:results}

\subsection{A Denotation Threshold Between 1.7B and 4B}
\label{sec:results:main}

Table~\ref{tab:main_results} reports the full ablation: 5 methods,
4 splits, 4 Qwen3 scales. On the eight sub-threshold cells (0.6B
and 1.7B), ReAct
reaches the highest compositional accuracy ($28.7$--$31.1$\,pp
above the next-best method) and dominates on every other split.
At 4B and above the picture inverts: free-form CoT wins
six of eight cells outright, and grammar-constrained EV+CoT wins
the remaining two (both at 8B on in-distribution and
cross-species). The threshold sits between 1.7B and 4B.

The species-stratified cross-proteome shift (mammal, avian,
invertebrate) does not flip any winner relative to in-distribution,
and hard-negative changes non-agentic accuracy by less than three
points uniformly (some splits up, some down). The denotation
threshold therefore reflects model capability, not a quirk of the
human-only test pool.

\begin{table*}[t]
\centering\scriptsize
\setlength{\tabcolsep}{3pt}
\renewcommand{\arraystretch}{0.85}
\begin{tabular}{ll ccc ccc c}
\toprule
\textbf{Model} & \textbf{Split}
  & \textbf{Standard} & \textbf{CoT} & \textbf{\scriptsize $\Delta$CoT}
  & \textbf{EV} & \textbf{EV+CoT} & \textbf{\scriptsize $\Delta$EV+CoT}
  & \textbf{ReAct} \\
\midrule
\multicolumn{9}{l}{\small\textsc{Qwen3-0.6B}} \\[-4pt]
\midrule
  & in-distribution & 21.15 & 20.21 & \textit{$-$0.94} & 15.55 & 14.22 & \textit{$-$1.33} & \textbf{27.62} \\
  & compositional & \phantom{0}4.93 & \phantom{0}3.40 & \textit{$-$1.53} & 12.49 & \phantom{0}6.34 & \textit{$-$6.15} & \textbf{43.57} \\
  & cross-species & 20.82 & 19.86 & \textit{$-$0.96} & 15.80 & 14.13 & \textit{$-$1.67} & \textbf{28.45} \\
  & HN & 19.90 & 19.14 & \textit{$-$0.76} & 17.93 & 15.71 & \textit{$-$2.22} & \textbf{29.15} \\
\midrule
\multicolumn{9}{l}{\small\textsc{Qwen3-1.7B}} \\[-4pt]
\midrule
  & in-distribution & 28.44 & 26.49 & \textit{$-$1.95} & 35.41 & 36.59 & $+$1.18 & \textbf{51.84} \\
  & compositional & \phantom{0}3.52 & \phantom{0}5.42 & $+$1.90 & 13.17 & \phantom{0}8.67 & \textit{$-$4.50} & \textbf{41.83} \\
  & cross-species & 28.24 & 26.79 & \textit{$-$1.45} & 34.45 & 36.02 & $+$1.57 & \textbf{52.91} \\
  & HN & 28.69 & 25.45 & \textit{$-$3.24} & 35.08 & 34.54 & \textit{$-$0.54} & \textbf{52.05} \\
\midrule
\multicolumn{9}{l}{\small\textsc{Qwen3-4B}} \\[-4pt]
\midrule
  & in-distribution & 69.62 & \textbf{81.60} & $+$11.98 & 70.01 & 81.41 & $+$11.40 & 60.01 \\
  & compositional & 46.07 & \textbf{87.57} & $+$41.50 & 27.57 & 72.09 & $+$44.52 & 33.85 \\
  & cross-species & 70.33 & \textbf{81.88} & $+$11.55 & 70.52 & 81.44 & $+$10.92 & 59.81 \\
  & HN & 70.32 & \textbf{83.13} & $+$12.81 & 69.71 & 82.06 & $+$12.35 & 60.36 \\
\midrule
\multicolumn{9}{l}{\small\textsc{Qwen3-8B}} \\[-4pt]
\midrule
  & in-distribution & 67.73 & 82.19 & $+$14.46 & 75.98 & \textbf{82.53} & $+$6.55 & 67.32 \\
  & compositional & 56.55 & \textbf{86.12} & $+$29.57 & 53.96 & 74.58 & $+$20.62 & 38.13 \\
  & cross-species & 67.29 & 81.29 & $+$14.00 & 76.08 & \textbf{82.21} & $+$6.13 & 68.15 \\
  & HN & 68.05 & \textbf{82.76} & $+$14.71 & 75.38 & 80.78 & $+$5.40 & 63.71 \\
\bottomrule
\end{tabular}
\caption{\textbf{Accuracy (\%) across five inference-time methods, four splits, and four Qwen3 scales.}
\textbf{Bold} = best per (model, split); \textit{italic} = negative $\Delta$.
Reasoning-prefix and grammar-planning effects:
$\Delta_{\mathrm{CoT}}(m,s) = A(m, \mathrm{CoT}, s) - A(m, \mathrm{Standard}, s)$ and
$\Delta_{\mathrm{EV+CoT}}(m,s) = A(m, \mathrm{EV+CoT}, s) - A(m, \mathrm{EV}, s)$,
where $A$ is the marginal accuracy at model $m$ on split $s$.
EV/EV+CoT: 3-seed mean. Others: seed 0.}
\label{tab:main_results}
\end{table*}

\begin{figure*}[t]
\centering
\begin{minipage}[t]{0.48\textwidth}
  \includegraphics[width=\linewidth]{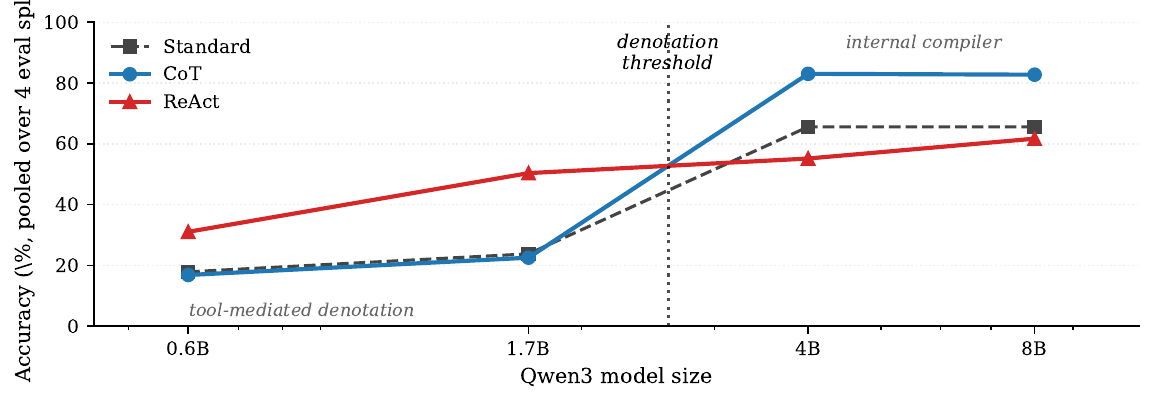}
  \subcaption{Denotation threshold (Qwen3).}
  \label{fig:threshold}
\end{minipage}%
\hfill
\begin{minipage}[t]{0.48\textwidth}
  \includegraphics[width=\linewidth]{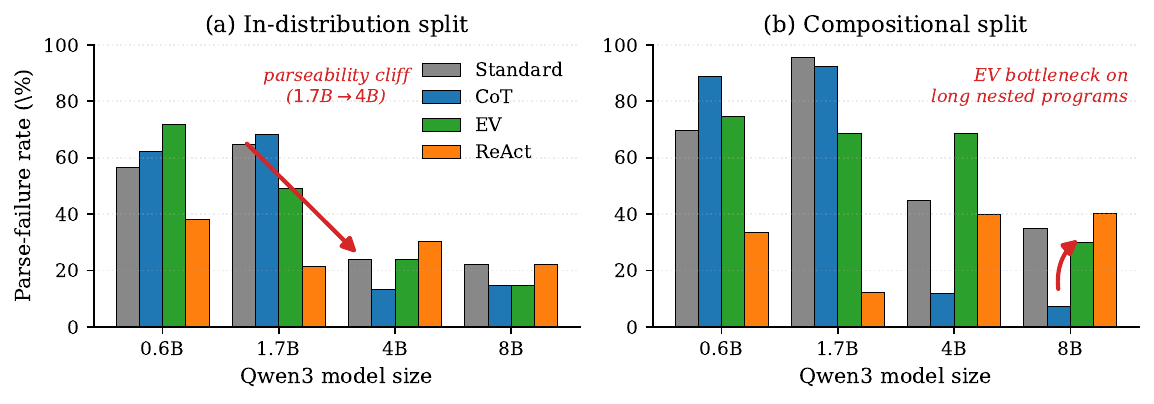}
  \subcaption{Parseability decomposition (Qwen3).}
  \label{fig:parseability}
\end{minipage}
\caption{\textbf{The denotation threshold and its parseability
mechanism.} (a) Pooled accuracy across all four splits for
Standard, CoT, and ReAct. (b) Parse-failure rate
($1{-}P(\mathrm{valid})$) per method on Qwen3. Per-cell
parseability rates for in-distribution and compositional splits in
App.~\ref{app:parse_failure}.}
\label{fig:denotation}
\end{figure*}

\subsection{The Threshold is a Parseability Transition}
\label{sec:results:crossover}

The denotation threshold (Figure~\ref{fig:denotation}) is
explained by a change in failure mode. Below threshold, small
models fail before semantics: they emit malformed programs,
invalid answer types, or free-form text that cannot be parsed
into the required denotation. Above threshold, parseability
improves sharply, and remaining errors are more often semantic
or aggregation errors.

The threshold becomes mechanistic when we decompose accuracy as
\[
  P(\mathrm{correct}) \;=\;
  P(\mathrm{valid}) \cdot
  P(\mathrm{correct}\mid\mathrm{valid}).
\]
$P(\mathrm{valid})$ is the rate at which the model produces a
parseable typed output, and $P(\mathrm{correct}\mid\mathrm{valid})$
the rate at which a parseable output matches the gold. Below the
threshold, models fail mainly on $P(\mathrm{valid})$; above, the
remaining headroom is in $P(\mathrm{correct}\mid\mathrm{valid})$,
which is what chain-of-thought improves.

Per-cell parse-failure rates (App.~\ref{app:parse_failure})
make this concrete: between Qwen3-1.7B and Qwen3-4B, Standard and
CoT parse-failure both drop sharply, and the pattern replicates
between Gemma-3-1B and Gemma-3-12B. At Qwen3-8B compositional,
grammar-constrained EV still has higher parse-failure than
free-form CoT.

The result is therefore not ``CoT always helps'' or ``grammar
always helps.'' Each method changes a different term in the
decomposition: tools and grammar improve $P(\mathrm{valid})$;
CoT improves semantic planning only after the model can maintain
a valid output channel; and constrained sampling can hurt long
compositional programs when syntactic validity narrows the
generation path too aggressively.

\subsection{Grammar is Selectively Valuable}
\label{sec:results:grammar}

Family-level results
(Table~\ref{tab:per_family_8b}, App.~\ref{app:per_family}) show
that grammar and execution remain valuable even above the
threshold, but not uniformly. Family C (PAE region-pair
aggregation) and Family E (secondary-structure labels) continue
to favor EV+CoT over free-form CoT at 8B because the correct
program must preserve exact region boundaries, pair orientation,
or discrete structural labels; deterministic execution prevents
the arithmetic and formatting errors that free-form CoT can
introduce.

In contrast, families that reduce to scalar lookup or arithmetic
over AlphaFold-derived quantities (Families A~(confidence),
B~(distance), D~(solvent exposure), F~(contact topology)) favor
free-form CoT at 4B and 8B. Once a model can identify the correct primitive
and scope, strict program generation becomes less necessary and
may introduce sampling failure.

The compositional split is the clearest stress test. CoT succeeds
when the model can reason through the operator composition in
text and emit the final value, while EV+CoT can fail because
generating a long nested grammar-valid program is itself
difficult. Failure under EV+CoT in this regime does not mean the
DSL cannot express the answer; it means the model struggles to
generate the correct executable expression under constraints.

\subsection{Agents as Denotation Prosthesis for Sub-Threshold Models}
\label{sec:results:react}

ReAct dominates at 0.6B and 1.7B because it supplies a
\emph{denotation prosthesis}: instead of requiring the model to
internally compile the full question into an executable
expression, tools expose local structural operations and raw DSL
execution. For sub-threshold models, this externalises the part of
the task they are least able to perform. At 4B and 8B, ReAct
loses this advantage: once models can internally compile most
questions, multi-turn tool use introduces over-deliberation,
formatting errors, and lost turn budget. 

ReAct's compositional accuracy is highest at the sub-threshold
scales, drops at $4$B, and partially recovers at $8$B. The 8B agent runs significantly longer
and over-constructs candidate programs that fail to execute, so
additional turns reduce rather than improve accuracy. We describe
this non-monotonic trajectory cautiously: the evidence supports a
minimum at 4B and a mechanism involving over-deliberation, but
not a precise functional form across model scale. The 8B failure
is sub-family-localized, concentrated on G1 and G3 (full
decomposition in App.~\ref{app:errors}).

\subsection{Replication on Gemma-3}
\label{sec:results:gemma3}

The same flip replicates on Gemma-3-1B/12B~\citep{gemma3} (ReAct
wins at $1$B, CoT wins at $12$B), confirming the threshold is not
Qwen3-specific (App.~\ref{app:gemma3_detail},
Table~\ref{tab:xfam_gemma3}).

\section{Conclusion}
\label{sec:conclusion}

Scientific QA should not stop at plausible answers. In structural
biology, the meaning of a question is a measurement over a
molecular object. ProtStructQA makes that measurement executable
by pairing natural-language questions with hidden typed programs
and deterministic answers over AlphaFold-predicted structures.
This executable view reveals a denotation threshold: small models
need tools to reach the structure, while larger models can use
chain-of-thought as an internal compiler from words to structural
measurements. By reframing protein QA as compilation from language
to denotation, ProtStructQA provides both a benchmark and a
diagnostic lens for scientific language grounding in the AlphaFold
era. ProtStructQA highlights executable grounding as a useful
direction for evaluating scientific QA beyond plausible answer
generation.

\section{Limitations}
\label{sec:limitations}

ProtStructQA tests structural reasoning under generated
natural-language questions, not the full distribution of questions
asked by practicing structural biologists. Although templates have
multiple paraphrases and family-specific lexical variation, the
language remains controlled. Families A--F share template identity
between train and test (different proteins and parameter values),
so the in-distribution split measures parameter-OOD, not
template-OOD; Family G provides the only template-OOD test.
Broader human-authored language generalization is left to future
work.

Gold answers are correct relative to AlphaFold-predicted structures
and the per-residue annotations derived from them, not relative to
experimental crystal structures or dynamic biology. This is
intentional: the benchmark evaluates whether models can query a
specified predicted structural object, and low-confidence regions
are kept to preserve a realistic uncertainty signal. ProtStructQA
results should not be interpreted as validating AlphaFold
predictions or as experimental claims about the underlying
proteins.

The denotation threshold is observed across Qwen3-$0.6$B to
Qwen3-$8$B and replicated on Gemma-3-$1$B and Gemma-3-$12$B
(\S\ref{sec:results:gemma3}). We do not claim that the
parameter-count location of the threshold is universal across
model families, training recipes, or larger scales. The
cross-family replication is limited to two Gemma scales and three
baselines (Standard, CoT, ReAct), so broader cross-family
evaluation, including more model families, more intermediate sizes,
and the full five-baseline suite, is an important next step.
Fine-tuning is out of scope.

The DSL covers single-chain monomeric structural measurements over
coordinates, confidence, PAE, solvent exposure, secondary
structure, and contacts. It does not model dynamics, ligand
binding, allostery, multimeric assemblies, evolutionary covariation,
or experimental uncertainty beyond AlphaFold-derived confidence
quantities.

The non-monotonic ReAct-on-compositional trajectory in
\S\ref{sec:results:react} rests on three Qwen3 data points;
resolving the curve shape would require intermediate scales. We
therefore describe the pattern as ``non-monotonic with a minimum
at $4$B'' and interpret the mechanism (tool use helps sub-threshold
models, but can introduce over-deliberation and formatting failures
above threshold) rather than claiming a universal scaling law.

\section{Ethics Statement}
\label{sec:ethics}

ProtStructQA is constructed from publicly released AlphaFold
Database predictions~\citep{jumper2021alphafold,bertoni2025afdb}
(CC~BY~4.0) and UniProt~\citep{uniprot2025} reference proteomes,
with attribution and licensing aligned to the upstream resources.
The benchmark contains no human subjects or personally identifying
information. Code is released under the MIT license; the derived
dataset is released under CC~BY~4.0, matching the upstream
UniProt and AFDB licenses.

The benchmark evaluates question answering over existing predicted
structures and does not provide design objectives for optimizing
pathogenicity, toxicity, immune evasion, or experimental
protocols. Protein-analysis tools can nevertheless be dual-use in
broader settings, so we avoid framing the benchmark as a
protein-design capability and preserve standard biosecurity
caveats.

Because answers are relative to predicted structures, users should
not treat benchmark outputs as experimental biological facts. The
intended use is evaluation of structural QA and
language-to-denotation reasoning, not wet-lab decision making.

\bibliography{custom}

\appendix

\section{Qualitative Evidence for the Denotation Threshold}
\label{app:errors}

\begin{figure*}[t]
\centering
\includegraphics[width=0.95\textwidth]{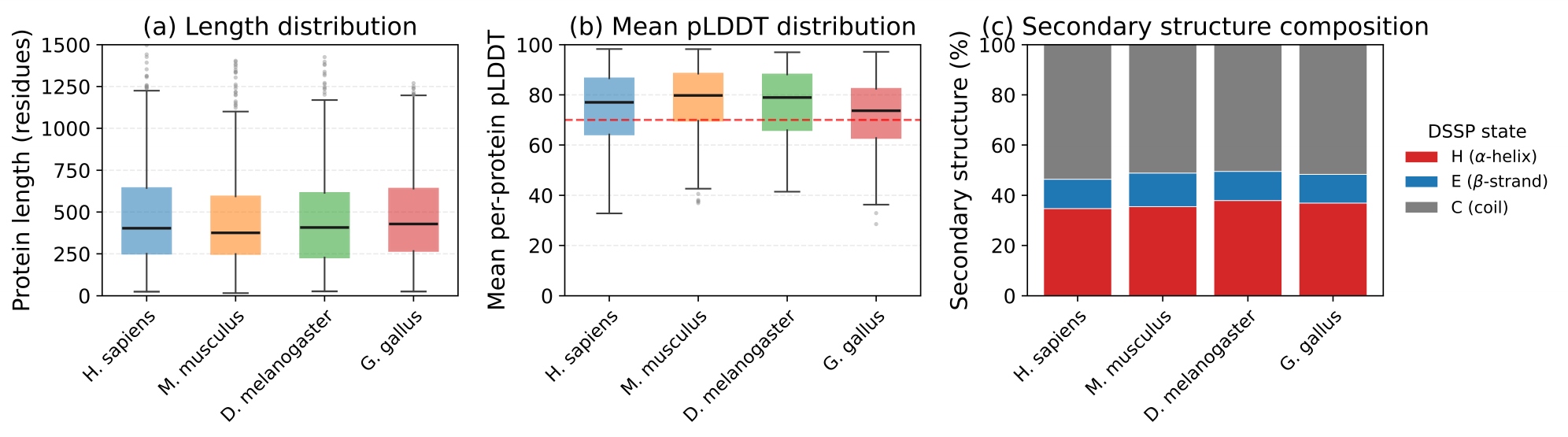}
\caption{\textbf{Protein-panel composition.}
(a)~per-species length distribution;
(b)~per-protein mean pLDDT;
(c)~DSSP H/E/C composition.}
\label{fig:protein_panel_dist}
\end{figure*}

\begin{table}[t]
\centering\footnotesize
\setlength{\tabcolsep}{3pt}
\begin{tabular}{@{}lrl@{}}
\toprule
Species & Proteins & Role \\
\midrule
Human (\textit{H.~sapiens})        & 4{,}000 & in-distribution anchor \\
Mouse (\textit{M.~musculus})       & 2{,}500 & mammalian cross-proteome \\
Fly (\textit{D.~melanogaster})     & 1{,}500 & invertebrate cross-proteome \\
Chicken (\textit{G.~gallus})       & 2{,}000 & avian cross-proteome \\
\midrule
\textbf{Total}                     & \textbf{10{,}000} & \\
\bottomrule
\end{tabular}
\caption{\textbf{Protein panel.} Human anchors in-distribution; mouse,
chicken, and fly define cross-proteome shifts.}
\label{tab:protein_panel}
\end{table}

\paragraph{ReAct 8B sub-family failure decomposition.}
At Qwen3-8B, ReAct fails almost entirely on the compositional
sub-families G1 (set-valued, ResidueSet, $\approx$13\%
accuracy) and G3 (nested-quantifier Boolean, $\approx$0\%
accuracy), while reaching near-ceiling on the simpler Boolean
sub-family G2 ($\approx$100\%). The two failure modes differ.
G1 failures are formatting-driven: inspection of failing rollouts
shows the agent identifies the correct residues during the tool
loop but commits its final answer as free-form prose rather than
a typed ResidueSet literal, so the deterministic executor never
sees the answer (Example~4 below). G3 failures are not a
formatting issue (the answer type is already Boolean); rather,
the agent rarely completes the nested
existential-with-distance check correctly within the available
turn budget. Together these effects explain the 36-point 8B
compositional gap between ReAct and EV+CoT.

We present four concrete examples drawn directly from the per-question
output logs. The first three illustrate failure modes that aggregate
into the denotation-threshold behavior reported in
Section~\ref{sec:results}; the
fourth illustrates the ReAct agent's answer-formatting failure mode
on Family-G1 questions (Section~\ref{sec:results:react}).

\paragraph{Example 1 (1.7B): Grammar+vote rescues a small model from
ambiguous-question misinterpretation.}
\begin{quote}
Question: ``Average AlphaFold pLDDT, 75 to 130?''\\
Family A, type Float, gold: $80.86$
\end{quote}
\textbf{Failure mode (Standard/CoT): validity}. The emitted program
cannot be parsed.
The phrase ``75 to 130'' is naturally read as a residue range
(\texttt{range(75, 130)}). The free-form 1.7B model misinterprets
``75'' as a pLDDT \emph{threshold}:
\begin{itemize}\setlength\itemsep{0pt}
  \item Standard 1.7B emits the malformed program
        \texttt{mean\_plddt(range(1,187) where plddt(r) >= 75 and plddt(r) <= 130)}
        and a free-form ``Answer: 82.6''; neither parses, so
        \texttt{pred=None}. $\times$
  \item CoT 1.7B makes the same misinterpretation
        (\texttt{mean\_plddt(all\_residues) where plddt(r) >= 75 and plddt(r) <= 130}).
        $\times$
  \item EV 1.7B samples three grammar-constrained programs;
        the consensus is \texttt{mean\_plddt(range(75, 130))} which
        executes to $80.86$. $\checkmark$
\end{itemize}
Grammar constraints prevent the small model from emitting
syntactically near-valid but semantically wrong programs, and the
$k{=}3$ vote breaks ties when one of the three samples is correct.

\paragraph{Example 2 (8B compositional): Free-form CoT handles
longer nested logic that grammar-constrained sampling fails to
generate reliably.}
\begin{quote}
Question: ``Is at least one 40-residue stretch high-confidence
(mean pLDDT $>70$) AND contact-rich (contact density $>0.3$)?''\\
Family G, type Bool, gold: \texttt{False}
\end{quote}
\textbf{Failure mode (EV+CoT): validity}. Grammar-constrained
sampling cannot emit a parseable nested program.
This is a Family G compositional question requiring (i) a sliding
window over 40-residue stretches, (ii) two predicate evaluations per
window, and (iii) an existential aggregator.
\begin{itemize}\setlength\itemsep{0pt}
  \item CoT 8B reasons step-by-step in free text and commits
        to \texttt{False}. $\checkmark$
  \item EV+CoT 8B fails to generate the required longer nested DSL
        expression under grammar-constrained sampling; all $k{=}3$
        samples emit unparseable or partial programs and
        \texttt{pred=None}. $\times$
\end{itemize}
This pattern, where grammar-constrained sampling fails on longer
nested expressions that free-form CoT simulates in text, accounts
for the $\sim$11.5pp gap by which CoT beats EV+CoT on the
compositional split at 8B.

\paragraph{Example 3 (8B Family C): Grammar wins on region-pair
PAE aggregation even at large scale.}
\begin{quote}
Question: ``What is the average AlphaFold relative-position
uncertainty between the 333--369 and 380--433 regions?''\\
Family C, type Float, gold: $5.77$
\end{quote}
\textbf{Failure mode (CoT): denotation}. The free-form trace
identifies the right primitive but computes the aggregation
incorrectly.
This is a single-step PAE aggregation, but 8B free-form CoT
mis-arithmetics the result.
\begin{itemize}\setlength\itemsep{0pt}
  \item CoT 8B writes a multi-paragraph derivation and
        outputs $2.99$ (a wrong arithmetic mean). $\times$
  \item EV+CoT 8B samples
        \texttt{mean\_pae(range(333,369), range(380,433))}, executes
        it deterministically, and returns $5.77$. $\checkmark$
\end{itemize}
This explains why even at 8B, Family C (PAE) is dominated by EV+CoT
by 22.7pp (Table~\ref{tab:per_family_8b}): aggregations over
contiguous regions are simple to express in DSL but error-prone for
the model to compute in free-form text.

\paragraph{Example 4 (1.7B ReAct on G1):
Agent reasoning succeeds, answer formatting fails.}
\begin{quote}
Question (Family G1, type ResidueSet): ``Which residues are
simultaneously buried (rel SASA $<$ 0.2) AND poorly predicted
(pLDDT $<$ 70)?''\\
Gold: \texttt{[188]} \emph{(single-residue ResidueSet)}
\end{quote}
\textbf{Failure mode (ReAct): answer-format}. The agent's tool
observations identify the correct residue, but the final
\texttt{<answer>} commit is in free-form prose rather than a
typed ResidueSet literal.
The 1.7B ReAct agent runs five turns: it iteratively calls
\texttt{run\_dsl} and \texttt{inspect\_residue}, narrowing to a
candidate region beginning at residue 188.
But on the final turn it emits the free-form sentence:
\begin{quote}\itshape
``residues 188--293 have a mean pLDDT of 66.20\ldots\ all residues in
this region are simultaneously buried (rel SASA $<$ 0.2) AND poorly
predicted (pLDDT $<$ 70).''
\end{quote}
The parser cannot extract a ResidueSet literal from this string, so
\texttt{pred=None} and the answer is scored wrong. The agent's
reasoning identifies residue 188 (the gold answer) at the start of
its candidate region, but the final \texttt{<answer>} commit is in
natural language rather than the expected \texttt{[N,\ldots]} typed
format. This illustrates a structural limitation of multi-turn agents
on structured-output benchmarks: without the deterministic executor
in the loop on the final commit (as in EV / EV+CoT), agent-generated
answers can fail type-formatting checks even when the underlying
reasoning is partially correct.

\paragraph{Summary of failure modes.}
Across the per-question logs we observe four dominant patterns:
(1) at small scale (1.7B), free-form text misinterprets ambiguous
question fragments and fails to commit a parseable program; grammar
+ vote rescues it; (2) on compositional questions at any scale,
grammar-constrained sampling becomes less reliable on longer
nested expressions (an underlying generation difficulty, not a
DSL expressivity limit), while free-form CoT can reason in text
and emit only the final scalar; (3) on region-pair aggregations, free-form models produce
internally inconsistent arithmetic, while the executor delivers a
correct answer deterministically; (4) multi-turn ReAct agents reason
correctly through the tool loop but fail to format the final
\texttt{<answer>} when the expected output type is structurally
ambiguous (e.g., Family G1 ResidueSet answers). The
denotation-threshold finding is the aggregate of the first three
patterns weighted by family composition; the fourth characterizes
a distinct, complementary weakness of agentic methods.

\begin{figure*}[t]
\centering
\small
\setlength{\tabcolsep}{4pt}
\renewcommand{\arraystretch}{1.25}
\definecolor{famHL}{HTML}{F5F8FC}
\begin{tabular}{@{}>{\centering\arraybackslash}p{0.8cm}
                  >{\raggedright\arraybackslash}p{2.0cm}
                  >{\raggedright\arraybackslash}p{1.9cm}
                  >{\raggedright\arraybackslash}p{3.8cm}
                  >{\raggedright\arraybackslash}p{3.7cm}
                  >{\raggedright\arraybackslash}p{1.4cm}@{}}
\toprule
\textbf{Family} & \textbf{Domain (type)} & \textbf{Protein} & \textbf{Natural-language question} & \textbf{Gold DSL program} & \textbf{Answer} \\
\midrule
\rowcolor{famHL}
A & Confidence (Float) &
  \texttt{Hu/P20155} &
  \textit{``What is the mean pLDDT of residues 14 to 55?''} &
  \texttt{mean\_plddt(range(14,55))} &
  $80.69$ \\
B & Distance (Float) &
  \texttt{Mo/Q9DCP9} &
  \textit{``What $C_\alpha$--$C_\alpha$ distance separates residues 73 and 57?''} &
  \texttt{distance(residue(73),} \texttt{\ residue(57))} &
  $15.2$\,\AA{} \\
\rowcolor{famHL}
B & Distance (PairSet) &
  \texttt{Hu/A6NHN6} &
  \textit{``Return pairs $(i,j)$ with $|i-j| > 20$ and distance$(i,j) < 10$ \AA{}.''} &
  \texttt{filter (i,j) in all\_pairs(} \texttt{\ min\_sep=20) where} \texttt{\ distance(i,j) < 10} &
  $\{(9,75),$ $(12,72), (12,75),$ $\ldots\}$ \emph{(11 pairs)} \\
C & PAE (Float) &
  \texttt{Hu/A6NJY4} &
  \textit{``What is the average AlphaFold relative-position uncertainty between the 14--30 and 41--70 regions?''} &
  \texttt{mean\_pae(range(14,30),} \texttt{\ range(41,70))} &
  $4.13$ \\
\rowcolor{famHL}
D & SASA (Bool) &
  \texttt{Hu/Q9NTN3} &
  \textit{``Is residue 46 hidden from solvent (rel SASA $<$ 0.2) in this structure?''} &
  \texttt{rel\_sasa(residue(46))} \texttt{< 0.2} &
  True \\
E & Secondary structure (SecStruct) &
  \texttt{Dm/Q6IHK7} &
  \textit{``Residue 49: H/E/C?''} &
  \texttt{ss(residue(49))} &
  H \\
\rowcolor{famHL}
F & Topology (Float) &
  \texttt{Gg/Q5ZJ39} &
  \textit{``What is the local packing density (contact fraction) for residues 37--56?''} &
  \texttt{contact\_density(} \texttt{\ range(37,56))} &
  $0.353$ \\
G & Compositional (Bool) &
  \texttt{Gg/A0A8V0ZRP1} &
  \textit{``Can a 40-residue window with mean\_plddt $>70$ AND contact\_density $>0.2$ be found?''} &
  \texttt{exists reg in sliding\_window(40)} \texttt{where mean\_plddt(reg) > 70} \texttt{and contact\_density(reg) > 0.2} &
  False \\
\bottomrule
\end{tabular}
\caption{\textbf{Example questions across the seven question families,
drawn from the paper-eval splits.}
Each row shows one question, the AlphaFold protein that grounds it
(\texttt{Hu}=human, \texttt{Mo}=mouse, \texttt{Dm}=fly,
\texttt{Gg}=chicken), the gold DSL program, and the gold answer. At
evaluation time, the model receives the question and a compact
structural summary of the protein. Family B is shown twice to
illustrate that answers can be compact (single Float) or non-compact
(a PairSet of residue pairs); other non-compact answer types
include Region and ResidueSet (\S\ref{sec:method}). Rows A--D are
from the in-distribution split, E--F from the cross-species split,
and G from the compositional split
(\S\ref{sec:benchmark:splits}).}
\label{fig:examples}
\end{figure*}

\section{Per-Family Accuracy at 8B}
\label{app:per_family}

Even at 8B where CoT wins overall, family-level inversion persists
on the structured-output facets. Families C (PAE) and E (secondary
structure) favor the grammar+execution branch (EV+CoT) by $22.7$
and $6.4$\,pp respectively over CoT, while Families A (pLDDT
aggregates), B (distance comparisons), D (SASA aggregates), and F
(contact topology) favor CoT, whose answers reduce to small
arithmetic comparisons the free-form chain can simulate directly.
The per-family pattern shows that grammar constraints are not
globally beneficial or globally harmful: their value depends on
whether exact typed structure prevents the dominant error mode for
that family.

\begin{table*}[t]
\centering\small
\begin{tabular}{l rrrr}
\toprule
Family & Standard & EV & EV+CoT & CoT \\
\midrule
A (Confidence)        & $63.8$ & $77.5$ & $86.7$ & $\textbf{93.2}$ \\
B (Distance)          & $79.4$ & $82.5$ & $84.4$ & $\textbf{88.1}$ \\
C (PAE)               & $52.2$ & $73.8$ & $\textbf{75.4}$ & $52.7$ \\
D (SASA)              & $76.6$ & $71.9$ & $81.2$ & $\textbf{87.2}$ \\
E (Secondary structure) & $57.9$ & $70.3$ & $\textbf{83.7}$ & $77.3$ \\
F (Topology)          & $68.0$ & $77.2$ & $82.3$ & $\textbf{86.2}$ \\
\bottomrule
\end{tabular}
\caption{\textbf{Per-family accuracy (\%) on the in-distribution
split at 8B.} Bold = best per family. Interpretation in the
section preamble.}
\label{tab:per_family_8b}
\end{table*}

\section{Hard-Negative Sub-class Breakdown}
\label{app:hn_breakdown}

The hard-negative split combines two construction types.
\textbf{HN1} (structural counterfactual, $n{=}4{,}138$, $95\%$)
keeps the template and protein fixed and resamples template
parameters until the gold answer changes. \textbf{HN2} (threshold
flip, $n{=}219$, $5\%$) keeps the protein, template, and
residues fixed and changes only a numeric threshold.
Table~\ref{tab:hn_breakdown} reports the per-class accuracy for
all five baselines at each scale.

HN2 is uniformly harder than HN1: all 20 (model, baseline) cells
satisfy HN1${>}$HN2, with mean gap $+10.3$\,pp. The widest gaps
occur for ReAct at supra-threshold scales (4B: $+18.9$\,pp; 8B:
$+22.9$\,pp), indicating that the agent's tool-call routine
over-relies on the specific threshold value seen in the query.
EV+CoT at 4B exhibits the narrowest gap ($+2.9$\,pp): grammar
enforcement closes the threshold-sensitivity gap once the base
model is fluent in the DSL. The HN2 sample size is small
($n{=}219$ per (model, baseline); some per-template strata
contain only $\approx 44$ examples), so per-cell standard errors
are roughly $\pm 7$\,pp at $95\%$ confidence.

\begin{table}[t]
\centering\small
\setlength{\tabcolsep}{3.5pt}
\begin{tabular}{ll rrrrr}
\toprule
Model & Class & Standard & CoT & EV & EV+CoT & ReAct \\
\midrule
\multirow{3}{*}{\textsc{0.6B}}
  & HN1 & $20.4$ & $20.0$ & $18.6$ & $16.4$ & $\mathbf{29.6}$ \\
  & HN2 & $11.0$ & $\phantom{0}3.7$ & $\phantom{0}5.2$ & $\phantom{0}3.2$ & $\mathbf{20.6}$ \\
  & $\Delta$ & $+9.4$ & $+16.3$ & $+13.4$ & $+13.2$ & $+9.1$ \\
\midrule
\multirow{3}{*}{\textsc{1.7B}}
  & HN1 & $29.0$ & $25.8$ & $35.5$ & $35.0$ & $\mathbf{52.4}$ \\
  & HN2 & $23.7$ & $18.7$ & $27.3$ & $26.8$ & $\mathbf{45.2}$ \\
  & $\Delta$ & $+5.2$ & $+7.1$ & $+8.3$ & $+8.2$ & $+7.2$ \\
\midrule
\multirow{3}{*}{\textsc{4B}}
  & HN1 & $70.9$ & $\mathbf{83.5}$ & $70.5$ & $82.2$ & $61.4$ \\
  & HN2 & $60.3$ & $77.2$ & $55.0$ & $\mathbf{79.3}$ & $42.5$ \\
  & $\Delta$ & $+10.6$ & $+6.3$ & $+15.6$ & $+2.9$ & $+18.9$ \\
\midrule
\multirow{3}{*}{\textsc{8B}}
  & HN1 & $68.6$ & $\mathbf{83.2}$ & $75.8$ & $81.0$ & $64.9$ \\
  & HN2 & $58.0$ & $\mathbf{75.3}$ & $67.1$ & $76.9$ & $42.0$ \\
  & $\Delta$ & $+10.6$ & $+7.8$ & $+8.7$ & $+4.1$ & $+22.9$ \\
\bottomrule
\end{tabular}
\caption{\textbf{Per-class accuracy (\%) on the hard-negative
split.} Bold = best per (model, class) row. HN1/HN2 defined in the
section preamble; $\Delta$ = HN1${-}$HN2. Interpretation in the
section preamble.}
\label{tab:hn_breakdown}
\end{table}

\section{Gold-Answer Correctness Validation}
\label{app:validation}

We validate gold correctness at two levels, complementing the 6-gate
sanity validator (\S\ref{sec:benchmark:validation}):
(i) \emph{paraphrase fidelity} between the natural-language question
and the gold DSL program, and (ii) \emph{cross-tool agreement} between
our pipeline's per-primitive values and an independent reference
toolchain. Paraphrase fidelity is run on $n{=}500$ per family
($n{=}3{,}500$ total); cross-tool agreement uses per-template
subsamples detailed below.

\paragraph{Paraphrase fidelity.}
For every sampled question, we verify (a)~that every numeric literal
in the gold DSL program (residue indices, range endpoints, thresholds)
also appears in the paraphrase, and (b)~that the paraphrase mentions
at least one family-specific concept term from a hand-curated lexicon
(e.g., ``pLDDT'' for A; ``distance'' for B; ``PAE'' for C; ``SASA''
or ``neighbors'' for D; ``helix'' or ``DSSP'' for E; ``contact'' or
``radius of gyration'' for F). On $3{,}500$ sampled questions
($500$ per family), $3{,}500/3{,}500$ ($100\%$) pass both checks.
The check is designed to catch obvious paraphrase/program
mismatches (numeric drift, missing referents, wrong family concept)
rather than to prove full semantic equivalence between every
paraphrase and its canonical program.

\paragraph{Cross-tool agreement.}
For each family that maps to a standard biophysical primitive, we
re-compute the underlying quantity with an independent reference and
compare to the value our pipeline stores. Implementations differ in
code path; in some cases (e.g., SASA) the algorithm itself differs.

\begin{itemize}\setlength\itemsep{0pt}
  \item \textbf{Family A (pLDDT).} \emph{Reference:} BioPython
        \texttt{PDBParser} reading the B-factor column from the
        AlphaFold PDB. Our pipeline uses a hand-written fixed-column
        parser. \emph{Agreement:} $500/500$ questions match the
        BioPython recomputation; max abs.\ difference
        $1.5 \times 10^{-5}$ pLDDT, mean $3.5 \times 10^{-6}$
        (float-precision round-trip).
  \item \textbf{Family B (distance).} \emph{Reference:} BioPython
        CA-atom coordinates piped through \texttt{numpy.linalg.norm}.
        Our DSL operates on the cached \texttt{ca\_xyz} array.
        \emph{Agreement:} $249/249$ (template B1, distance float);
        max abs.\ difference $6.0 \times 10^{-6}$\,\AA{}.
  \item \textbf{Family C (PAE).} \emph{Reference:} AlphaFold per-pair
        PAE JSON loaded with stdlib \texttt{json}. AFDB delivers PAE
        as integer values, which our pipeline stores as a uint8
        matrix (lossless re-encoding); agreement is measured after
        dequantization to float. \emph{Agreement:} $133/133$ (template
        C1, mean PAE over inter-region block); max abs.\ difference
        $9.4 \times 10^{-7}$ (float-precision noise).
  \item \textbf{Family D (SASA / neighbor counts).} SASA is
        validated against BioPython \texttt{Bio.PDB.SASA.ShrakeRupley}
        (pure-Python Shrake--Rupley) versus our pipeline's FreeSASA C
        library~\citep{mitternacht2016freesasa} (different algorithm):
        per-residue Spearman $\rho = 0.995$ (mean over $104$ proteins,
        $\rho_{\min} = 0.987$); buried/exposed classification at
        threshold-far residues: $96/96$ match. Neighbor counts are
        deterministic $C_\alpha$-distance computations and are
        independently recomputed from coordinates in the validation
        script.
  \item \textbf{Family E (DSSP H/E/C).} Our extractor uses
        \texttt{pydssp} (pure-Python DSSP~\citep{kabsch1983dssp},
        H-bond based). \emph{Reference:}
        biotite's native P-SEA secondary-structure annotation
        \citep{labesse1997psea}, which classifies SS from
        $C_\alpha$-trace geometric criteria (inter-residue distances
        and dihedral angles) rather than backbone H-bonds
        (algorithmically independent). Across $273$ proteins,
        $94.2\%$ per-residue agreement on helix-vs-non-helix and
        $79.6\%$ 3-state agreement (DSSP and P-SEA disagree at
        $\alpha$-helix capping boundaries by construction). Gold-answer
        match: $273/273$ on templates E1, E2, E3.
  \item \textbf{Family F (contacts and topology).}
        Contact-density templates reduce to $C_\alpha$-coordinate
        distance computations and inherit Family B agreement; the
        remaining geometric aggregates (e.g., radius of gyration,
        contact count) are deterministic coordinate functions
        independently recomputed in the validation script and matched
        within float precision.
  \item \textbf{Family G (compositional)} inherits the primitive-level
        checks of A--F (each compositional gold answer is produced by
        composing already-validated A--F primitives).
\end{itemize}

\section{Stratified Evaluation Subsampling Protocol}
\label{app:subsample}

All four paper-eval splits are released. The three subsamples
(cross-species, compositional, hard-negative) are stratified by
template$\times$species (cross-species and compositional) and by
template (hard-negative). They are rigorously train-disjoint, and
the distribution shift is bounded by KL$\,\leq 0.0037$ on every
(family, species, template, answer-type) marginal (see manifest
in the release), so subsample trends transfer to the full pool
within statistical noise. Split sizes and pool sizes are listed
in \S\ref{sec:benchmark:subsample}.

\begin{table*}[t]
\centering\footnotesize
\setlength{\tabcolsep}{6pt}
\begin{tabular}{@{}l r l@{}}
\toprule
Split & Released & Role \\
\midrule
Train (A--F)               &  96{,}000 & Prompt-example pool \\
Dev (A--F)                 &  12{,}000 & Prompt dev / sanity checks \\
In-distribution test       &  12{,}000 & Eval (parameter generalization) \\
Compositional (Family G)   &  30{,}000 & Eval (held-out compositions) \\
Cross-species (A--F)       & 180{,}000 & Eval (cross-proteome shift) \\
Hard-negative robustness   &  52{,}200 & Robustness probe \\
\midrule
\textbf{Total released}    & \textbf{382{,}200} & \\
\midrule
\textbf{Paper evaluation subsample} & \textbf{32{,}357}
  & Stratified eval subset \\
\bottomrule
\end{tabular}
\caption{\textbf{Full split inventory.} The released benchmark
contains $382{,}200$ questions across six splits. The paper
evaluates models on a $32{,}357$-question stratified subsample
drawn from the four evaluation splits: $12$K in-distribution,
$6$K compositional, $10$K cross-species, and $4{,}357$ HN.}
\label{tab:split_inventory}
\end{table*}

\section{Family and Template Question Distribution}
\label{app:per_template}

Table~\ref{tab:families} summarises the full $382{,}200$-question
release at the family level: each family contributes between $3$ and
$6$ templates and roughly $30$K--$60$K questions across the active
benchmark, with the $52.2$K hard-negative robustness pool counted on
its own row.

\begin{table}[t]
\centering\scriptsize
\setlength{\tabcolsep}{2.5pt}
\begin{tabular}{@{}llrr@{}}
\toprule
Family & Domain & \#Templates & \#Questions \\
\midrule
A & pLDDT (confidence)        & 5 & 59{,}617 \\
B & Distance                  & 4 & 60{,}145 \\
C & PAE                       & 4 & 43{,}437 \\
D & SASA / packing            & 5 & 58{,}814 \\
E & Secondary structure       & 6 & 35{,}112 \\
F & Topology / contacts       & 4 & 42{,}875 \\
G & \emph{Compositional}      & 3 & 30{,}000 \\
\midrule
HN robustness pool          & ---     & --- & 52{,}200 \\
\midrule
\textbf{Total} & & \textbf{31} & \textbf{382{,}200} \\
\bottomrule
\end{tabular}
\caption{\textbf{Template families and per-family question counts.}
Family G (held out) and the HN robustness pool are on their own
rows.}
\label{tab:families}
\end{table}

Table~\ref{tab:per_template} breaks this down at the template level,
restricted to the four evaluation pools (in-distribution test, full
compositional pool, full cross-species pool, full hard-negative pool;
$274{,}200$ questions in total). Templates B1--B4 (Distance) and
G1--G3 (Compositional) carry the largest mass (10K--13K each), while
the secondary-structure templates E3--E6 are smaller (3.2K--3.4K)
because they target rare topology patterns. Counts are determined by
the per-protein sampling budget at question-generation time
(\S\ref{sec:benchmark}).

\begin{table*}[!t]
\centering\scriptsize
\setlength{\tabcolsep}{4pt}
\begin{tabular}{lr|lr|lr|lr|lr|lr|lr}
\toprule
\multicolumn{2}{c|}{Family A} & \multicolumn{2}{c|}{Family B} &
\multicolumn{2}{c|}{Family C} & \multicolumn{2}{c|}{Family D} &
\multicolumn{2}{c|}{Family E} & \multicolumn{2}{c|}{Family F} &
\multicolumn{2}{c}{Family G} \\
Template & Count & Template & Count & Template & Count & Template & Count &
Template & Count & Template & Count & Template & Count \\
\midrule
A1 & 10{,}291 & B1 & 12{,}724 & C1 &  8{,}974 & D1 &  9{,}661 & E1 &  6{,}050 & F1 &  8{,}941 & G1 & 11{,}443 \\
A2 &  9{,}986 & B2 & 10{,}890 & C2 &  8{,}793 & D2 &  9{,}174 & E2 &  5{,}992 & F2 &  9{,}066 & G2 & 12{,}025 \\
A3 &  9{,}745 & B3 & 12{,}376 & C3 &  9{,}173 & D3 &  9{,}352 & E3 &  3{,}297 & F3 &  9{,}116 & G3 &  9{,}951 \\
A4 &  9{,}640 & B4 & 12{,}281 & C4 &  8{,}891 & D4 & 10{,}245 & E4 &  3{,}353 & F4 &  8{,}422 &    &          \\
A5 &  8{,}083 &    &          &    &          & D5 &  9{,}759 & E5 &  3{,}220 &    &          &    &          \\
   &          &    &          &    &          &    &          & E6 &  3{,}286 &    &          &    &          \\
\bottomrule
\end{tabular}
\caption{\textbf{Per-template question counts across the four full released
evaluation pools.} In-distribution test (12K), full compositional pool
(30K), full cross-species pool (180K), and full hard-negative
robustness pool (52.2K). Each column corresponds to one family A--G;
empty cells in shorter families are blank. The 96K train and 12K dev
exemplar pools are not shown; together with these four pools they
bring the full released benchmark to $382{,}200$ questions
(Table~\ref{tab:split_inventory}). Templates with smaller counts
(E3--E6) target rare secondary-structure topologies and are sampled
at lower frequency to reflect their natural distribution. Subtotal of
the four evaluation pools shown above: $274{,}200$ questions (of the
$382{,}200$-question release; $96{,}000$ train and $12{,}000$ dev are
not shown in this table). Because the hard-negative pool is generated
from the same templates, per-family totals in this table include
HN-derived questions and therefore differ from the active-benchmark
per-family counts in Table~\ref{tab:families} (e.g., Family G is
$30{,}000$ in Table~\ref{tab:families} versus $33{,}419$ here).}
\label{tab:per_template}
\end{table*}

\section{DSL Vocabulary and Complete Template Catalogue}
\label{app:dsl_catalogue}

This appendix is intended to make the abstract description in
\S\ref{sec:benchmark} concrete. We list (i) every operator in the
DSL in plain English, and (ii) the exact program pattern
of every one of the 31 templates the benchmark uses.

\paragraph{What is the DSL?}
The DSL is a small formal language that lets us write a
short program describing a question. A program takes a single protein
as input and returns a single answer. The grammar is fixed, and the
executor runs each program deterministically against the protein's
extracted features (pLDDT, distances, PAE, SASA, secondary structure).

\paragraph{What is a template?}
A template is a parameterised question pattern. It bundles four
things: a canonical DSL program with named placeholder slots, the
expected answer type, the rules that decide which parameters are
legal for a given protein, and a list of natural-language
paraphrases that re-word the same question ($26$--$30$ for A--F,
$5$--$6$ for G). The
$31$ templates are author-designed. The paraphrase pool was generated
by an LLM (Claude Opus~4.7) and then author-verified to remove paraphrases
that revealed the answer and to make sure every paraphrase kept the
same slot names as its template. Every other step of the pipeline
(protein selection, parameter sampling, gold-program slot-filling,
DSL execution, paraphrase choice, and split assignment) is
deterministic given the released random seeds.

\paragraph{How a question is built.} For every protein we pick a
template, sample legal parameters, fill the template's canonical DSL
program with those parameters to obtain the gold program, execute it
to obtain the gold answer, and fill one randomly chosen paraphrase
to obtain the natural-language question text. At evaluation time,
the model sees the question text together with a compact protein
summary. For the non-agent baselines, the model must produce a valid
DSL program. For ReAct, the model can call DSL tools and then return
a final typed answer. Figure~\ref{fig:question_pipeline} walks
through this end-to-end for one concrete question.

\begin{figure}[t]
\centering\footnotesize
\newcommand{\pipestep}[2]{%
  \fbox{\begin{minipage}{0.93\columnwidth}\footnotesize
  \textbf{Step~#1.}~#2
  \end{minipage}}\\[1pt]
  $\downarrow$\\[1pt]%
}
\newcommand{\pipelast}[2]{%
  \fcolorbox{black}{green!4}{\begin{minipage}{0.93\columnwidth}\footnotesize
  \textbf{Step~#1.}~#2
  \end{minipage}}%
}
\pipestep{1.~Pick a protein}{%
  Source: UniProt reference proteomes. Picked:
  \texttt{Hu/O60637} (length $L\!=\!253$, AlphaFold-predicted).}
\pipestep{2.~Pick a template (one of $31$)}{%
  Picked: \texttt{A1\_RegionMeanPLDDT} (Family~A, answer type
  \texttt{Float}). Canonical DSL program:
  \texttt{mean\_plddt(range(\{start\}, \{end\}))}.}
\pipestep{3.~Sample legal parameters}{%
  A1 rule: $s\!\in\![1, L\!-\!W]$, window $W\!\in\![20, 80]$.
  Sampled: \texttt{\{start: 60, end: 121\}}.}
\pipestep{4.~Build gold DSL program (fill slots)}{%
  Program: \texttt{mean\_plddt(range(60, 121))}.}
\pipestep{5.~Execute (deterministic DSL executor)}{%
  Reads the protein's pLDDT array, computes
  $\mathrm{mean}(\mathrm{pLDDT}[59{:}121])\!=\!91.54$.
  Gold answer: $91.54$.}
\pipestep{6.~Pick one paraphrase (from A1's pool of $30$)}{%
  Seeded \texttt{random.choice} $\to$ \texttt{paraphrase\_id\,=\,12}.
  Picked: \emph{``Over positions \{start\}-\{end\}, what is the
  average per-residue pLDDT?''}}
\pipestep{7.~Fill the paraphrase slots}{%
  Final natural-language question, shown to the model alongside the
  compact protein summary: \emph{``Over positions 60-121, what is the
  average per-residue pLDDT?''}}
\fcolorbox{black}{green!4}{%
\begin{minipage}{0.93\columnwidth}\footnotesize
\textbf{Step~8.~Save one question record (one row in the released
JSONL file).}\par\vspace{2pt}
{\scriptsize\ttfamily
\{\\
\hspace*{1em}"qid":~~~~~~~~~~~"human/O60637/A1/0",\\
\hspace*{1em}"uniprot":~~~~~~~"O60637",\\
\hspace*{1em}"species":~~~~~~~"human",\\
\hspace*{1em}"family":~~~~~~~~"A",\\
\hspace*{1em}"template":~~~~~~"A1",\\
\hspace*{1em}"question":~~~~~~"Over positions 60-121, ...?",\\
\hspace*{1em}"program":~~~~~~~"mean\_plddt(range(60, 121))",\\
\hspace*{1em}"answer":~~~~~~~~91.54,\\
\hspace*{1em}"answer\_type":~~~"Float",\\
\hspace*{1em}"params":~~~~~~~~\{"start":~60, "end":~121\},\\
\hspace*{1em}"paraphrase\_id":~12\\
\}\par}
\end{minipage}}
\caption{\textbf{End-to-end pipeline for producing one ProtStructQA
question, traced for a concrete example.} Steps~1--5 produce the
gold program and gold answer (no natural language involved).
Steps~6--7 pick one of the template's paraphrases ($26$--$30$ for
A--F templates, $5$--$6$ for G templates) and fill the same slots
to obtain the natural-language question text. Step~8 shows the actual JSON record stored in the released
benchmark file. All steps are deterministic given fixed random
seeds; the entire $382{,}200$-question benchmark is reproducible
from the released paraphrase pool and seed configuration.}
\label{fig:question_pipeline}
\end{figure}

\subsection*{Plain-English DSL vocabulary}

\paragraph{Per-residue values (one number per residue).}
\begin{itemize}\setlength\itemsep{0pt}
  \item \texttt{plddt(r)}: AlphaFold's confidence for residue $r$
        (range $0$--$100$; higher is more confident).
  \item \texttt{ss(r)}: secondary structure of residue $r$ as a
        DSSP letter (\texttt{"H"} helix, \texttt{"E"} strand,
        \texttt{"C"} coil).
  \item \texttt{rel\_sasa(r)}: relative solvent-accessible surface
        area; values near $0$ mean buried, near $1$ mean exposed.
  \item \texttt{n\_neighbors(r)}: how many other residues sit within
        $8$\,\AA{} of residue $r$ (a packing measure).
\end{itemize}

\paragraph{Per-pair values (one number per pair of residues).}
\begin{itemize}\setlength\itemsep{0pt}
  \item \texttt{distance(r1, r2)}: $C_\alpha$--$C_\alpha$ distance
        in \AA{}.
  \item \texttt{pae(r1, r2)}: predicted aligned error in \AA{} for
        the relative position of two residues (lower is more reliable).
\end{itemize}

\paragraph{Region-level aggregates (one number per contiguous span).}
\begin{itemize}\setlength\itemsep{0pt}
  \item \texttt{mean\_plddt(reg)}, \texttt{min\_plddt(reg)},
        \texttt{max\_plddt(reg)}: summary pLDDT over a span.
  \item \texttt{mean\_rel\_sasa(reg)}: average accessibility in a
        span.
  \item \texttt{contact\_density(reg)}: fraction of in-span residue
        pairs that are in contact ($<$$8$\,\AA{}).
  \item \texttt{radius\_of\_gyration(reg)}: how compact the span is.
  \item \texttt{mean\_pae(reg1, reg2)}, \texttt{max\_pae(reg1, reg2)},
        \texttt{count\_high\_pae(reg1, reg2, $\tau$)}: inter-region
        PAE summaries.
\end{itemize}

\paragraph{Whole-protein scalars.}
\texttt{n\_helices()}, \texttt{n\_strands()},
\texttt{length(longest\_run("H"))}.

\paragraph{Constructors (how to refer to residues or spans).}
\texttt{residue(i)}: a single residue. \texttt{range(s, e)}: the
contiguous span from $s$ to $e$. \texttt{first(k)} / \texttt{last(k)}
-- the first / last $k$ residues. \texttt{sliding\_window(k)}: the
set of all length-$k$ contiguous spans. \texttt{all\_residues}: the
set of every residue. \texttt{all\_pairs(min\_sep=k)}: the set of
ordered residue pairs $(i, j)$ with sequence separation
$|i - j| > k$.

\paragraph{Comprehensions and operators (how to combine the above).}
\texttt{count} / \texttt{filter} / \texttt{exists} / \texttt{forall}
$x$ \texttt{in} $S$ \texttt{where} $P$: iterate $x$ over $S$ and
either count, collect, or test the predicate $P$.
\texttt{argmin} / \texttt{argmax} $x$ \texttt{in} $S$ \texttt{by} $E$
-- return the element of $S$ that minimizes / maximizes the
expression $E$. Comparison operators ($<$, $\le$, $==$, $\ne$, $>$,
$\ge$) and Boolean operators (\texttt{and}, \texttt{or},
\texttt{not}) compose predicates.

\subsection*{Complete catalogue of the 31 templates}

Table~\ref{tab:template_catalogue} lists every template, the
canonical DSL program it produces (with named slots), the answer
type, and one example paraphrase. Slot names match the variables
the parameter sampler fills in. In template~A2, the program slot
\texttt{w} is a local alias for the parameter \texttt{window}, and
\texttt{term1} is the terminus (\texttt{"N"} or \texttt{"C"})
sampled per question; the program shown in the table is the
\texttt{term1="N"} branch. Family A--F templates are seen during
training as few-shot exemplars; Family G is held out, so the
compositional split tests genuinely unseen template types. The
hard-negative split is generated by perturbing parameters within
the existing $31$ templates.

\begin{table*}[t]
\centering\footnotesize
\setlength{\tabcolsep}{4pt}
\begin{tabular}{@{}l l l p{6.4cm} p{5.2cm}@{}}
\toprule
ID & Fam. & Answer type & Canonical DSL program & Example paraphrase \\
\midrule
A1 & A & Float & \texttt{mean\_plddt(range(\{start\}, \{end\}))} & What is the mean pLDDT of residues \{start\} to \{end\}? \\
A2 & A & Bool & \texttt{mean\_plddt(first(\{window\})) < mean\_plddt(last(\{window\}))} & Across the first and last \{window\} residues, is the \{term1\}-terminal pLDDT lower? \\
A3 & A & Region & \texttt{argmin reg in sliding\_window(\{window\}) by mean\_plddt(reg)} & Which \{window\}-residue window has the lowest mean pLDDT? \\
A4 & A & Int & \texttt{count r in all\_residues where plddt(r) > \{threshold\}} & How many residues have pLDDT > \{threshold\}? \\
A5 & A & Bool & \texttt{exists reg in sliding\_window(\{window\}) where mean\_plddt(reg) > \{threshold\}} & Does this protein have a \{window\}-residue region with mean pLDDT above \{threshold\}? \\
\midrule
B1 & B & Float & \texttt{distance(residue(\{i\}), residue(\{j\}))} & Distance between residues \{i\} and \{j\}? \\
B2 & B & Bool & \texttt{distance(residue(\{i\}), residue(\{j\})) < \{threshold\}} & Residues \{i\}, \{j\} within \{threshold\} Angstroms? \\
B3 & B & PairSet & \texttt{filter (i,j) in all\_pairs(min\_sep=\{sep\}) where distance(i,j) < \{threshold\}} & Pairs with sep > \{sep\} and CA-CA < \{threshold\} Angstroms? \\
B4 & B & Int & \texttt{size(filter (i,j) in all\_pairs(min\_sep=\{sep\}) where distance(i,j) < \{threshold\})} & How many sep>\{sep\}, d<\{threshold\} Angstroms pairs? \\
\midrule
C1 & C & Float & \texttt{mean\_pae(range(\{a\_start\}, \{a\_end\}), range(\{b\_start\}, \{b\_end\}))} & Mean PAE, (\{a\_start\}-\{a\_end\}) vs (\{b\_start\}-\{b\_end\})? \\
C2 & C & Bool & \texttt{mean\_pae(range(\{a\_start\},\{a\_end\}), range(\{b\_start\},\{b\_end\})) < \{threshold\}} & Boolean: mean inter-region PAE for (\{a\_start\}-\{a\_end\}, \{b\_start\}-\{b\_end\}) < \{threshold\}? \\
C3 & C & Float & \texttt{max\_pae(range(\{a\_start\},\{a\_end\}), range(\{b\_start\},\{b\_end\}))} & Max PAE, (\{a\_start\}-\{a\_end\}) vs (\{b\_start\}-\{b\_end\})? \\
C4 & C & Int & \texttt{count\_high\_pae(range(\{a\_start\},\{a\_end\}), range(\{b\_start\},\{b\_end\}), \{threshold\})} & Count of high-PAE pairs (> \{threshold\}), block (\{a\_start\}-\{a\_end\}) by (\{b\_start\}-\{b\_end\})? \\
\midrule
D1 & D & Bool & \texttt{rel\_sasa(residue(\{i\})) < \{threshold\}} & Is residue \{i\}'s relative SASA below \{threshold\}? \\
D2 & D & Region & \texttt{argmax reg in sliding\_window(\{window\}) by mean\_rel\_sasa(reg)} & Which \{window\}-residue window has the highest average solvent accessibility? \\
D3 & D & Int & \texttt{count r in all\_residues where rel\_sasa(r) < \{threshold\}} & How many residues have relative SASA < \{threshold\}? \\
D4 & D & Int & \texttt{n\_neighbors(residue(\{i\}))} & Count the residues within 8 A of residue \{i\}. \\
D5 & D & Bool & \texttt{n\_neighbors(residue(\{i\})) > \{threshold\}} & Is residue \{i\}'s 8-A neighbor count above \{threshold\}? \\
\midrule
E1 & E & SecStruct & \texttt{ss(residue(\{i\}))} & What is the secondary structure at residue \{i\}? \\
E2 & E & Bool & \texttt{ss(residue(\{i\})) == "H"} & Is residue \{i\} part of an alpha-helix? \\
E3 & E & Int & \texttt{count r in all\_residues where ss(r) == "H"} & How many residues are in alpha-helices? \\
E4 & E & Int & \texttt{count r in all\_residues where ss(r) == "E"} & How many residues are in beta-strands? \\
E5 & E & Int & \texttt{length(longest\_run("H"))} & How long is the longest helical segment? \\
E6 & E & Int & \texttt{n\_helices()} & Number of helix segments? \\
\midrule
F1 & F & Float & \texttt{contact\_density(range(\{start\}, \{end\}))} & What is the contact density of residues \{start\}-\{end\}? \\
F2 & F & Bool & \texttt{exists reg in sliding\_window(\{window\}) where mean\_plddt(reg) > 80 and contact\_density(reg) > \{cd\_thr\}} & Is there any \{window\}-window that is both mean pLDDT > 80 and contact density > \{cd\_thr\}? \\
F3 & F & Float & \texttt{radius\_of\_gyration(range(\{start\}, \{end\}))} & Rg for residues \{start\}-\{end\}? \\
F4 & F & Region & \texttt{argmin reg in sliding\_window(\{window\}) by radius\_of\_gyration(reg)} & Which \{window\}-residue window has the smallest radius of gyration (most compact)? \\
\midrule
G1 & G & ResidueSet & \texttt{filter r in all\_residues where rel\_sasa(r) < \{sasa\_thr\} and plddt(r) < \{plddt\_thr\}} & List residues that satisfy both rel\_sasa(r) < \{sasa\_thr\} and plddt(r) < \{plddt\_thr\}. \\
G2 & G & Bool & \texttt{exists reg in sliding\_window(\{window\}) where mean\_plddt(reg) > \{plddt\_thr\} and contact\_density(reg) > \{cd\_thr\}} & Is there a \{window\}-residue region with mean pLDDT > \{plddt\_thr\} AND contact density > \{cd\_thr\}? \\
G3 & G & Bool & \texttt{exists r in all\_residues where ss(r)=="H" and exists s in all\_residues where ss(s)=="E" and distance(r,s) < \{threshold\}} & Does any helix residue come within \{threshold\} A of a strand residue? \\
\bottomrule
\end{tabular}
\caption{\textbf{Complete catalogue of the 31 ProtStructQA
templates.} Each row shows the canonical DSL program (named
slots), required answer type, and one of $26$--$30$ paraphrases
(A--F templates; $5$--$6$ for G).
Notation (\texttt{w}, \texttt{term1}) and the held-out / HN
conventions are described in App.~\ref{app:dsl_catalogue}
preamble.}
\label{tab:template_catalogue}
\end{table*}

\section{Prompt-Form Sensitivity at Sub-Threshold}
\label{app:prompt_form}

To check whether the sub-threshold CoT result depends on our specific
prompt design, we evaluate two alternative CoT formulations at
Qwen3-1.7B alongside our task-decomposition variant:
\emph{vanilla zero-shot CoT}~\citep{kojima2022zeroshotcot}
(``Think step by step before answering''), and
\emph{PAL-style CoT}~\citep{gao2023pal,chen2023program}, where reasoning
is carried by inline ``\#''-comment annotations within the few-shot
exemplars rather than by a prompt prefix. Pooled accuracy across
all $32{,}357$ paper-eval questions at $1.7$B:

\begin{itemize}\setlength\itemsep{1pt}
  \item No CoT (Standard): $23.79\%$
  \item Vanilla zero-shot CoT (Kojima): $18.62\%$ (worst)
  \item Task-decomposition CoT (our published variant): $22.54\%$
  \item PAL-style CoT (exemplar comments): $24.47\%$ (best)
\end{itemize}

The three CoT variants span only $5.9$pp from worst to best; even
the best CoT prompt (PAL, $24.5\%$) sits $6.5$pp below EV ($31.0\%$)
and $25.8$pp below ReAct ($50.3\%$). Instruction-driven CoT
(task-decomposition and Kojima) does not improve over no-CoT direct
prompting at $1.7$B; PAL-style CoT delivers only a marginal lift over
no-CoT ($\Delta{=}+0.68$pp pooled; $p\!<\!10^{-5}$ on test\_iid by
McNemar).
Prompt-engineering perturbations cannot close the gap to the structural
methods that EV and ReAct realize; the sub-threshold capability
ceiling is robust to prompt-form choice.

\section{Statistical Significance Detail}
\label{app:stats_detail}

EV and EV+CoT use three seeds; per-cell seed noise is much smaller
than the threshold-defining accuracy gaps.
We report per-cell McNemar's paired-test results for all $80$
planned pairwise comparisons (5 method-pairs $\times$ 4 splits
$\times$ 4 model scales) with Bonferroni correction at
$\alpha = 0.05/80 = 6.25 \times 10^{-4}$.
Because McNemar's test requires matched per-question predictions
while Table~\ref{tab:main_results} reports EV and EV+CoT as
three-seed means, the paired statistics in
Tables~\ref{tab:significance_sub} and~\ref{tab:significance_supra}
use the seed-0 EV and EV+CoT runs. Some $\Delta$ values therefore
differ from the three-seed means in Table~\ref{tab:main_results}
by less than one accuracy point. Table~\ref{tab:significance_sub}
covers the sub-threshold scales (Qwen3-0.6B and 1.7B) and
Table~\ref{tab:significance_supra} covers the supra-threshold
scales (Qwen3-4B and 8B).

\textbf{34 of 40 sub-threshold} comparisons are significant; the
6 ties cluster where small models fail similarly or where EV and
EV+CoT are close at 1.7B. \textbf{31 of 40 supra-threshold}
comparisons are significant; the 9 ties cluster at CoT vs EV+CoT
and Standard vs EV at $\ge 4$B, where grammar adds little once the
base model is capable enough. Columns in both tables follow the
same convention: ``A-only''/``B-only'' = questions correct in
exactly one method of the pair; $\Delta$ accuracy $=$ (B$-$A),
so a positive $\Delta$ favors method B and a negative $\Delta$
favors method A; the
final row per (model, split) cell compares ReAct to the strongest
non-agentic baseline.

\begin{table*}[t]
\centering\small
\setlength{\tabcolsep}{4pt}
\begin{tabular}{ll lr rrl}
\toprule
Model & Split & Pair (A vs B) & $\Delta$ (pp) & A-only & B-only & McNemar $p$ \\
\midrule
\multirow{20}{*}{\textsc{Qwen3-0.6B}}
 & \multirow{5}{*}{in-distribution} & Standard vs CoT         & $-0.94$ & $833$ & $720$ & $2.6{\times}10^{-3}$ \,n.s. \\
 &  & EV vs EV{+}CoT     & $-1.32$ & $574$ & $415$ & $4.8{\times}10^{-7}$ $^\star$ \\
 &  & CoT vs EV{+}CoT    & $-6.16$ & $1{,}277$ & $538$ & $3.8{\times}10^{-69}$ $^\star$ \\
 &  & Standard vs EV          & $-5.78$ & $1{,}283$ & $590$ & $6.9{\times}10^{-59}$ $^\star$ \\
 &  & ReAct vs Standard       & $-6.47$ & $2{,}445$ & $1{,}669$ & $8.4{\times}10^{-34}$ $^\star$ \\
\cmidrule(lr){2-7}
 & \multirow{5}{*}{compositional} & Standard vs CoT         & $-1.53$ & $265$ & $173$ & $1.3{\times}10^{-5}$ $^\star$ \\
 &  & EV vs EV{+}CoT     & $-6.67$ & $615$ & $215$ & $2.2{\times}10^{-45}$ $^\star$ \\
 &  & CoT vs EV{+}CoT    & $+2.78$ & $175$ & $342$ & $1.8{\times}10^{-13}$ $^\star$ \\
 &  & Standard vs EV          & $+7.92$ & $207$ & $682$ & $7.4{\times}10^{-60}$ $^\star$ \\
 &  & ReAct vs EV        & $-30.72$ & $1{,}980$ & $137$ & $<10^{-300}$ $^\star$ \\
\cmidrule(lr){2-7}
 & \multirow{5}{*}{cross-species} & Standard vs CoT         & $-0.96$ & $656$ & $560$ & $9.3{\times}10^{-3}$ \,n.s. \\
 &  & EV vs EV{+}CoT     & $-1.47$ & $485$ & $338$ & $3.4{\times}10^{-7}$ $^\star$ \\
 &  & CoT vs EV{+}CoT    & $-5.70$ & $1{,}024$ & $454$ & $8.2{\times}10^{-51}$ $^\star$ \\
 &  & Standard vs EV          & $-5.19$ & $1{,}043$ & $524$ & $7.8{\times}10^{-40}$ $^\star$ \\
 &  & ReAct vs Standard       & $-7.63$ & $2{,}144$ & $1{,}381$ & $5.5{\times}10^{-38}$ $^\star$ \\
\cmidrule(lr){2-7}
 & \multirow{5}{*}{hard-negative} & Standard vs CoT         & $-0.76$ & $261$ & $228$ & $0.15$ \,n.s. \\
 &  & EV vs EV{+}CoT     & $-2.16$ & $261$ & $167$ & $6.4{\times}10^{-6}$ $^\star$ \\
 &  & CoT vs EV{+}CoT    & $-3.10$ & $368$ & $233$ & $4.1{\times}10^{-8}$ $^\star$ \\
 &  & Standard vs EV          & $-1.70$ & $386$ & $312$ & $5.0{\times}10^{-3}$ \,n.s. \\
 &  & ReAct vs Standard       & $-9.25$ & $924$ & $521$ & $1.8{\times}10^{-26}$ $^\star$ \\
\midrule
\multirow{20}{*}{\textsc{Qwen3-1.7B}}
 & \multirow{5}{*}{in-distribution} & Standard vs CoT         & $-1.95$ & $1,174$ & $940$ & $3.9{\times}10^{-7}$ $^\star$ \\
 &  & EV vs EV{+}CoT     & $+0.55$ & $991$ & $1,057$ & $0.15$ \,n.s. \\
 &  & CoT vs EV{+}CoT    & $+10.17$ & $840$ & $2,060$ & $5.6{\times}10^{-117}$ $^\star$ \\
 &  & Standard vs EV          & $+7.67$ & $754$ & $1,674$ & $1.5{\times}10^{-79}$ $^\star$ \\
 &  & ReAct vs EV{+}CoT  & $-15.18$ & $3{,}885$ & $2{,}063$ & $3.2{\times}10^{-125}$ $^\star$ \\
\cmidrule(lr){2-7}
 & \multirow{5}{*}{compositional} & Standard vs CoT         & $+1.90$ & $190$ & $304$ & $3.3{\times}10^{-7}$ $^\star$ \\
 &  & EV vs EV{+}CoT     & $-4.92$ & $517$ & $222$ & $5.6{\times}10^{-28}$ $^\star$ \\
 &  & CoT vs EV{+}CoT    & $+3.02$ & $268$ & $449$ & $1.4{\times}10^{-11}$ $^\star$ \\
 &  & Standard vs EV          & $+9.83$ & $204$ & $794$ & $1.0{\times}10^{-82}$ $^\star$ \\
 &  & ReAct vs EV        & $-28.48$ & $1{,}945$ & $236$ & $<10^{-300}$ $^\star$ \\
\cmidrule(lr){2-7}
 & \multirow{5}{*}{cross-species} & Standard vs CoT         & $-1.45$ & $920$ & $775$ & $4.7{\times}10^{-4}$ $^\star$ \\
 &  & EV vs EV{+}CoT     & $+1.53$ & $738$ & $891$ & $1.6{\times}10^{-4}$ $^\star$ \\
 &  & CoT vs EV{+}CoT    & $+9.11$ & $677$ & $1,588$ & $8.7{\times}10^{-84}$ $^\star$ \\
 &  & Standard vs EV          & $+6.13$ & $655$ & $1,268$ & $5.2{\times}10^{-45}$ $^\star$ \\
 &  & ReAct vs EV{+}CoT  & $-17.01$ & $3{,}348$ & $1{,}647$ & $2.2{\times}10^{-130}$ $^\star$ \\
\cmidrule(lr){2-7}
 & \multirow{5}{*}{hard-negative} & Standard vs CoT         & $-3.24$ & $446$ & $305$ & $3.0{\times}10^{-7}$ $^\star$ \\
 &  & EV vs EV{+}CoT     & $-0.78$ & $386$ & $352$ & $0.22$ \,n.s. \\
 &  & CoT vs EV{+}CoT    & $+9.43$ & $290$ & $701$ & $6.6{\times}10^{-40}$ $^\star$ \\
 &  & Standard vs EV          & $+6.98$ & $272$ & $576$ & $7.0{\times}10^{-26}$ $^\star$ \\
 &  & ReAct vs EV        & $-16.39$ & $1{,}374$ & $660$ & $1.8{\times}10^{-57}$ $^\star$ \\
\bottomrule
\end{tabular}
\caption{\textbf{McNemar paired-test results at sub-threshold
scales (0.6B and 1.7B), 40 comparisons.} Bonferroni
$\alpha = 6.25{\times}10^{-4}$; $^\star$ = significant.
Column conventions and significance summary in the section preamble.}
\label{tab:significance_sub}
\end{table*}

\begin{table*}[t]
\centering\small
\setlength{\tabcolsep}{4pt}
\begin{tabular}{ll lr rrl}
\toprule
Model & Split & Pair (A vs B) & $\Delta$ (pp) & A-only & B-only & McNemar $p$ \\
\midrule
\multirow{20}{*}{\textsc{Qwen3-4B}}
 & \multirow{5}{*}{in-distribution} & Standard vs CoT         & $+11.97$ & $487$ & $1,924$ & $6.2{\times}10^{-201}$ $^\star$ \\
 &  & EV vs EV{+}CoT     & $+11.80$ & $419$ & $1,835$ & $2.0{\times}10^{-210}$ $^\star$ \\
 &  & CoT vs EV{+}CoT    & $+0.48$ & $727$ & $785$ & $0.14$ \,n.s. \\
 &  & Standard vs EV          & $+0.66$ & $1,044$ & $1,123$ & $0.09$ \,n.s. \\
 &  & ReAct vs EV{+}CoT  & $+22.07$ & $513$ & $3{,}162$ & $<10^{-300}$ $^\star$ \\
\cmidrule(lr){2-7}
 & \multirow{5}{*}{compositional} & Standard vs CoT         & $+41.50$ & $142$ & $2,632$ & $<10^{-300}$ $^\star$ \\
 &  & EV vs EV{+}CoT     & $+45.08$ & $125$ & $2,830$ & $<10^{-300}$ $^\star$ \\
 &  & CoT vs EV{+}CoT    & $-15.22$ & $1,159$ & $246$ & $8.8{\times}10^{-142}$ $^\star$ \\
 &  & Standard vs EV          & $-18.80$ & $1,646$ & $518$ & $3.7{\times}10^{-136}$ $^\star$ \\
 &  & ReAct vs CoT       & $+53.72$ & $0$ & $3,223$ & $<10^{-300}$ $^\star$ \\
\cmidrule(lr){2-7}
 & \multirow{5}{*}{cross-species} & Standard vs CoT         & $+11.55$ & $371$ & $1,526$ & $7.6{\times}10^{-166}$ $^\star$ \\
 &  & EV vs EV{+}CoT     & $+10.99$ & $372$ & $1,471$ & $3.6{\times}10^{-154}$ $^\star$ \\
 &  & CoT vs EV{+}CoT    & $-0.38$ & $658$ & $620$ & $0.30$ \,n.s. \\
 &  & Standard vs EV          & $+0.18$ & $903$ & $921$ & $0.69$ \,n.s. \\
 &  & ReAct vs CoT       & $+22.07$ & $348$ & $2{,}555$ & $<10^{-300}$ $^\star$ \\
\cmidrule(lr){2-7}
 & \multirow{5}{*}{hard-negative} & Standard vs CoT         & $+12.81$ & $162$ & $720$ & $1.3{\times}10^{-84}$ $^\star$ \\
 &  & EV vs EV{+}CoT     & $+12.67$ & $141$ & $693$ & $3.1{\times}10^{-88}$ $^\star$ \\
 &  & CoT vs EV{+}CoT    & $-0.60$ & $270$ & $244$ & $0.27$ \,n.s. \\
 &  & Standard vs EV          & $-0.46$ & $426$ & $406$ & $0.51$ \,n.s. \\
 &  & ReAct vs CoT       & $+22.77$ & $130$ & $1{,}122$ & $2.1{\times}10^{-197}$ $^\star$ \\
\midrule
\multirow{20}{*}{\textsc{Qwen3-8B}}
 & \multirow{5}{*}{in-distribution} & Standard vs CoT         & $+14.47$ & $681$ & $2,417$ & $5.1{\times}10^{-226}$ $^\star$ \\
 &  & EV vs EV{+}CoT     & $+5.25$ & $514$ & $1,144$ & $3.7{\times}10^{-55}$ $^\star$ \\
 &  & CoT vs EV{+}CoT    & $-0.19$ & $1,155$ & $1,132$ & $0.65$ \,n.s. \\
 &  & Standard vs EV          & $+9.02$ & $887$ & $1,970$ & $2.3{\times}10^{-93}$ $^\star$ \\
 &  & ReAct vs CoT       & $+14.88$ & $728$ & $2{,}513$ & $6.8{\times}10^{-228}$ $^\star$ \\
\cmidrule(lr){2-7}
 & \multirow{5}{*}{compositional} & Standard vs CoT         & $+29.57$ & $398$ & $2,172$ & $1.6{\times}10^{-294}$ $^\star$ \\
 &  & EV vs EV{+}CoT     & $+20.58$ & $307$ & $1,542$ & $1.5{\times}10^{-197}$ $^\star$ \\
 &  & CoT vs EV{+}CoT    & $-11.75$ & $1,151$ & $446$ & $7.8{\times}10^{-72}$ $^\star$ \\
 &  & Standard vs EV          & $-2.77$ & $1,237$ & $1,071$ & $5.9{\times}10^{-4}$ $^\star$ \\
 &  & ReAct vs CoT       & $+47.98$ & $504$ & $3{,}383$ & $<10^{-300}$ $^\star$ \\
\cmidrule(lr){2-7}
 & \multirow{5}{*}{cross-species} & Standard vs CoT         & $+14.00$ & $598$ & $1,998$ & $5.4{\times}10^{-175}$ $^\star$ \\
 &  & EV vs EV{+}CoT     & $+5.72$ & $376$ & $948$ & $2.7{\times}10^{-57}$ $^\star$ \\
 &  & CoT vs EV{+}CoT    & $+0.79$ & $901$ & $980$ & $0.07$ \,n.s. \\
 &  & Standard vs EV          & $+9.07$ & $715$ & $1,622$ & $2.1{\times}10^{-80}$ $^\star$ \\
 &  & ReAct vs EV{+}CoT  & $+13.93$ & $764$ & $2{,}157$ & $2.7{\times}10^{-152}$ $^\star$ \\
\cmidrule(lr){2-7}
 & \multirow{5}{*}{hard-negative} & Standard vs CoT         & $+14.71$ & $246$ & $887$ & $2.0{\times}10^{-85}$ $^\star$ \\
 &  & EV vs EV{+}CoT     & $+5.10$ & $183$ & $405$ & $2.8{\times}10^{-20}$ $^\star$ \\
 &  & CoT vs EV{+}CoT    & $-2.23$ & $504$ & $407$ & $1.5{\times}10^{-3}$ \,n.s. \\
 &  & Standard vs EV          & $+7.39$ & $377$ & $699$ & $6.2{\times}10^{-23}$ $^\star$ \\
 &  & ReAct vs CoT       & $+19.05$ & $217$ & $1{,}047$ & $1.2{\times}10^{-130}$ $^\star$ \\
\bottomrule
\end{tabular}
\caption{\textbf{McNemar paired-test results at supra-threshold
scales (4B and 8B), 40 comparisons.} Bonferroni
$\alpha = 6.25{\times}10^{-4}$; $^\star$ = significant. Column
conventions and significance summary in the section preamble
(shared with Table~\ref{tab:significance_sub}).}
\label{tab:significance_supra}
\end{table*}

\section{Bootstrap Confidence Intervals}
\label{app:bootstrap_cis}

Table~\ref{tab:bootstrap_cis} reports the bootstrap $95\%$ CI
half-width for each (model, method, split) cell in
Table~\ref{tab:main_results}, computed by resampling question
indices $1{,}000$ times (EV/EV+CoT pool across the three seeds). Half-widths are typically below $1$\,pp on
in-distribution and cross-species cells, and reach $1.0$--$1.5$\,pp
on the smaller hard-negative split ($n{=}4{,}357$) and on ReAct
compositional cells where the agentic decoding has higher intrinsic
variance.

\begin{table}[t]
\centering\small
\setlength{\tabcolsep}{4pt}
\begin{tabular}{ll rrrrr}
\toprule
Model & Split & Std & CoT & EV & EV+CoT & ReAct \\
\midrule
\multirow{4}{*}{\textsc{0.6B}} & iid & 0.72 & 0.72 & 0.60 & 0.56 & 0.79 \\
 & comp & 0.57 & 0.46 & 0.62 & 0.41 & 1.24 \\
 & x-sp & 0.85 & 0.80 & 0.64 & 0.62 & 0.89 \\
 & HN & 1.21 & 1.22 & 1.02 & 0.92 & 1.30 \\
\midrule
\multirow{4}{*}{\textsc{1.7B}} & iid & 0.82 & 0.75 & 0.76 & 0.79 & 0.87 \\
 & comp & 0.45 & 0.58 & 0.75 & 0.56 & 1.28 \\
 & x-sp & 0.93 & 0.87 & 0.88 & 0.84 & 0.95 \\
 & HN & 1.39 & 1.25 & 1.38 & 1.23 & 1.43 \\
\midrule
\multirow{4}{*}{\textsc{4B}} & iid & 0.81 & 0.70 & 0.76 & 0.62 & 0.89 \\
 & comp & 1.28 & 0.85 & 0.88 & 1.04 & 1.15 \\
 & x-sp & 0.88 & 0.75 & 0.80 & 0.66 & 1.02 \\
 & HN & 1.31 & 1.04 & 1.19 & 0.98 & 1.48 \\
\midrule
\multirow{4}{*}{\textsc{8B}} & iid & 0.81 & 0.72 & 0.67 & 0.61 & 0.81 \\
 & comp & 1.20 & 0.83 & 1.13 & 0.93 & 1.20 \\
 & x-sp & 0.94 & 0.80 & 0.73 & 0.68 & 0.93 \\
 & HN & 1.32 & 1.11 & 1.12 & 1.08 & 1.42 \\
\bottomrule
\end{tabular}
\caption{\textbf{Per-cell bootstrap $95\%$ CI half-widths (pp).}
Full $95\%$ CI for any Table~\ref{tab:main_results} cell is point
estimate $\pm$ half-width. Split abbreviations: iid =
in-distribution ($n{=}12{,}000$), comp = compositional
($n{=}6{,}000$), x-sp = cross-species ($n{=}10{,}000$), HN =
hard-negative ($n{=}4{,}357$). Half-widths grow with smaller $n$
and with intrinsic decoding variance.}
\label{tab:bootstrap_cis}
\end{table}

\section{Parse-Failure Rate Decomposition}
\label{app:parse_failure}

To support the mechanism claims in
\S\ref{sec:results:crossover}, we measure the per-question
\emph{parse-failure rate} per (model, baseline) cell on the
in-distribution and compositional splits. A
parse failure is a question where the inference pipeline could not
produce a usable answer ($\texttt{pred\_answer} = $ None): for
Standard and CoT this means the free-form output yielded no
extractable DSL program; for EV and EV+CoT it means all $k{=}3$
grammar-constrained samples failed to execute; for ReAct it means
the agent never committed a parseable typed
\texttt{<answer>}~literal. Table~\ref{tab:parse_failure} reports the
overall rate by (model, baseline, split).

Two patterns support the \S\ref{sec:results:crossover} mechanism
claim. (1) Between $1.7$B and $4$B on the in-distribution split,
free-form parse-failure drops by $40.9$\,pp for Standard
($64.8\% {\to} 23.9\%$) and $54.9$\,pp for CoT
($68.4\% {\to} 13.4\%$); the capability threshold is in part a
\emph{parseability} threshold. (2) At $8$B on the compositional
split, free-form CoT reaches a $7.4\%$ parse-failure floor while
grammar-constrained EV remains at $29.9\%$ and EV+CoT at $23.3\%$,
consistent with grammar-constrained sampling becoming less
reliable on longer nested expressions.

\begin{table}[t]
\centering\footnotesize
\setlength{\tabcolsep}{3pt}
\begin{tabular}{ll ccccc}
\toprule
Model & Split & Standard & CoT & EV & EV+CoT & ReAct \\
\midrule
\multirow{2}{*}{\textsc{Qwen3-0.6B}}
  & in-distribution & 56.4 & 62.4 & 71.7 & 73.3 & 38.0 \\
  & compositional   & 69.6 & 89.0 & 74.5 & 85.2 & 33.3 \\
\midrule
\multirow{2}{*}{\textsc{Qwen3-1.7B}}
  & in-distribution & 64.8 & 68.4 & 48.9 & 55.7 & 21.5 \\
  & compositional   & 95.7 & 92.3 & 68.5 & 80.7 & 12.1 \\
\midrule
\multirow{2}{*}{\textsc{Qwen3-4B}}
  & in-distribution & 23.9 & 13.4 & 23.9 & 11.5 & 30.2 \\
  & compositional   & 44.9 & 11.8 & 68.6 & 25.4 & 39.9 \\
\midrule
\multirow{2}{*}{\textsc{Qwen3-8B}}
  & in-distribution & 22.1 & 14.6 & 14.7 & 12.1 & 22.1 \\
  & compositional   & 34.8 & \phantom{0}7.4 & 29.9 & 23.3 & 40.2 \\
\bottomrule
\end{tabular}
\caption{\textbf{Parse-failure rate (\%) by (model, baseline,
split).} $n{=}12{,}000$ for in-distribution; $n{=}6{,}000$ for
compositional. Interpretation in the section preamble.}
\label{tab:parse_failure}
\end{table}

\section{Cross-Family Replication: Gemma-3 Detail}
\label{app:gemma3_detail}

This appendix accompanies \S\ref{sec:results:gemma3} with the full
Gemma-3 per-cell accuracy.

\begin{table}[t]
\centering\small
\setlength{\tabcolsep}{6pt}
\begin{tabular}{l l rr}
\toprule
Model & Baseline & Overall & Compositional \\
\midrule
\multirow{3}{*}{Gemma-3-1B}
    & Standard &           12.73 &           2.37 \\
    & CoT      &           12.61 &           1.68 \\
    & ReAct    & \textbf{13.10}  & \textbf{8.65}  \\
\midrule
\multirow{3}{*}{Gemma-3-12B}
    & Standard &           67.37 &          51.55 \\
    & CoT      & \textbf{83.15}  & \textbf{75.67} \\
    & ReAct    &           65.56 &          41.97 \\
\bottomrule
\end{tabular}
\caption{\textbf{Replication on Gemma-3.} Accuracy (\%)
pooled across all four splits (Overall) and on compositional only.
\textbf{Bold} = winner per column. The 1B-ReAct $\to$ 12B-CoT flip
matches Qwen3 (Table~\ref{tab:main_results}).}
\label{tab:xfam_gemma3}
\end{table}

\end{document}